\documentclass{article}

\usepackage{microtype}
\usepackage{graphicx}
\usepackage{subfigure}
\usepackage{booktabs} 

\usepackage{hyperref}

\usepackage[accepted]{mlsys2022}

\usepackage{color}         
\usepackage{enumitem}

\usepackage{wrapfig}
\usepackage{algorithm}
\usepackage{algorithmic}
\usepackage{multirow}
\usepackage{graphicx}
\usepackage{xspace}
\usepackage{lipsum}
\usepackage{caption}
\usepackage{Definitions}

\usepackage{amsmath,amsfonts,bm}

\def\Figref#1{Figure~\ref{#1}}

\def\secref#1{section~\ref{#1}}
\def\Secref#1{Section~\ref{#1}}

\def\eqref#1{equation~\ref{#1}}

\def\1{\bm{1}}

\def\rb{{\textnormal{b}}}

\def\vb{{\bm{b}}}

\DeclareMathAlphabet{\mathsfit}{\encodingdefault}{\sfdefault}{m}{sl}
\SetMathAlphabet{\mathsfit}{bold}{\encodingdefault}{\sfdefault}{bx}{n}

\let\ab\allowbreak

\definecolor{darkpastelgreen}{rgb}{0.01, 0.75, 0.24}

\newcommand{\modelshort}{SMORE\xspace}
\newcommand{\modellong}{Scalable Multi-hOp REasoning \xspace}

\graphicspath{{./figs/}} 

\mlsystitlerunning{Scaling up Knowledge Graph Embeddings}

\begin{document}

\twocolumn[
\mlsystitle{\modelshort: Knowledge Graph Completion and Multi-hop Reasoning in Massive Knowledge Graphs}
\mlsyssetsymbol{equal}{*}

\begin{mlsysauthorlist}
\mlsysauthor{Hongyu Ren}{st,equal}
\mlsysauthor{Hanjun Dai}{goo,equal}
\mlsysauthor{Bo Dai}{goo}
\mlsysauthor{Xinyun Chen}{be}
\mlsysauthor{Denny Zhou}{goo}
\mlsysauthor{Jure Leskovec}{st}
\mlsysauthor{Dale Schuurmans}{goo}
\end{mlsysauthorlist}

\mlsysaffiliation{st}{Stanford University}
\mlsysaffiliation{goo}{Google Brain}
\mlsysaffiliation{be}{UC Berkeley}

\mlsyscorrespondingauthor{Hongyu Ren}{hyren@cs.stanford.edu}
\mlsyscorrespondingauthor{Hanjun Dai}{hadai@google.com}

\mlsyskeywords{Machine Learning, MLSys}

\vskip 0.3in
\begin{abstract}
Knowledge graphs (KGs) capture knowledge in the form of  {\em head--relation--tail} triples and are a crucial component in many AI systems. There are two important reasoning tasks on KGs: (1) single-hop knowledge graph completion, which involves predicting individual links in the KG; and (2), multi-hop reasoning, where the goal is to predict which KG entities satisfy a given logical query. Embedding-based methods solve both tasks by first computing an embedding for each entity and relation, then using them to form predictions. However, existing scalable KG embedding frameworks only support single-hop knowledge graph completion and cannot be applied to the more challenging multi-hop reasoning task. Here we present {\em \modellong (\modelshort)}, the first general framework for both single-hop and multi-hop reasoning in KGs. Using a \emph{single machine} \modelshort can perform multi-hop reasoning in Freebase KG (86M entities, 338M edges), which is 1,500$\times$ larger than previously considered KGs. The key to \modelshort's runtime performance is a novel bidirectional rejection sampling that achieves a square root reduction of the complexity of online training data generation. Furthermore, \modelshort exploits asynchronous scheduling,  overlapping CPU-based data sampling, GPU-based embedding computation, and frequent CPU--GPU IO. \modelshort increases throughput (\ie, training speed) over prior multi-hop KG frameworks by 2.2$\times$ with minimal GPU memory requirements (2GB for training 400-dim embeddings on 86M-node Freebase) and achieves near linear speed-up with the number of GPUs.  Moreover, on the simpler single-hop knowledge graph completion task \modelshort achieves comparable or even better runtime performance to state-of-the-art frameworks on both single GPU and multi-GPU settings.
\end{abstract}
]



\printAffiliationsAndNotice{\mlsysEqualContribution} 

\setlength{\abovedisplayskip}{2pt}
\setlength{\abovedisplayshortskip}{2pt}
\setlength{\belowdisplayskip}{2pt}
\setlength{\belowdisplayshortskip}{2pt}
\setlength{\jot}{1pt}

\setlength{\floatsep}{1ex}
\setlength{\textfloatsep}{1ex}

\section{Introduction}
A \emph{knowledge graph} (KG) is a heterogeneous graph structure that captures knowledge encoded in a form of {\em head--relation--tail} triples, where the {\em head} and {\em tail} are two entities (\ie, nodes) and the {\em relation} is an edge between them (\eg, {\em (Paris, CapitalOf, France)}).
Knowledge graphs form the backbone of many AI systems across a wide range of domains: recommender systems~\citep{wang2018dkn,ying2018graph,wang2019knowledge,wu2019dual}, question answering~\citep{berant2013semantic,sun2018open,sun2019pullnet,saxena2020improving,ren2021lego},  commonsense reasoning~\citep{lin2019kagnet,feng2020scalable,yasunaga2021qa,lv2020graph}, personalized medicine~\cite{ruiz2021identification} and drug discovery~\cite{zitnik2018modeling,drkg2020}.

\begin{figure}[t]
\centering
    \includegraphics[width=0.45\textwidth]{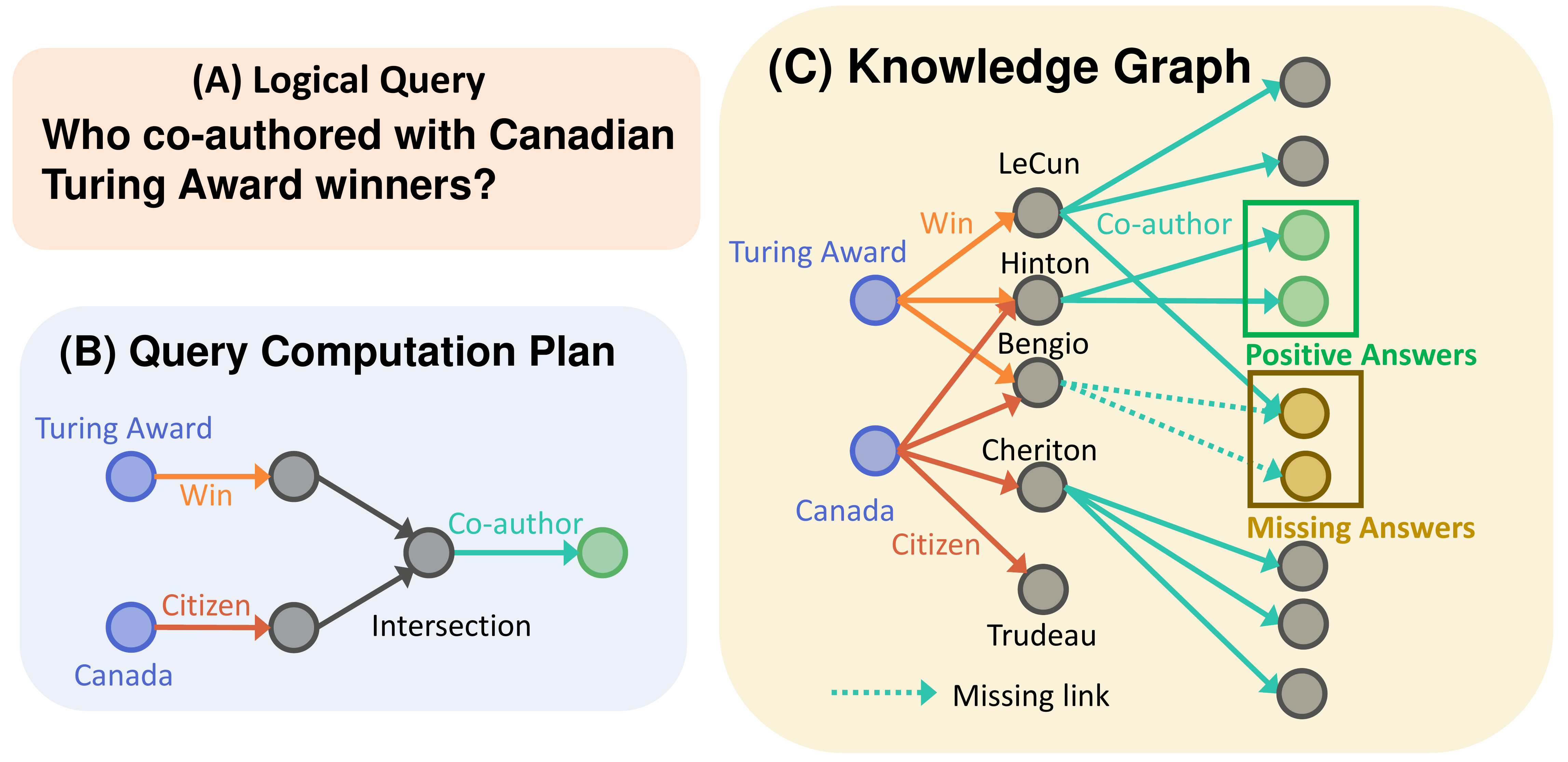}
    \caption{Query embedding methods aim to answer multi-hop logical queries (A) by avoiding explicit knowledge graph traversal and executing the query directly in the embedding space by following the query computation plan (B). Such methods are robust against missing links (C).
    }
    \vspace{-1mm}
\label{fig:background}
\end{figure}

Reasoning over such KGs consists of two types of tasks: (1) single-hop link prediction (also known as knowledge graph completion), where given a {\em head} and a {\em relation} the goal is to predict one or more {\em tail} entities. For example, given {\em TuringAward--Win--?} (\ie, Who are the Turing Award winners?), the goal is to predict entities {\em GeoffHinton, DonKnuth}, etc.; %
And, (2) multi-hop reasoning, where one needs to predict (one or many) of the {\em tails} of a multi-hop logical query. For example, answering ``Who are co-authors of Canadian Turing Award winners?'' (\Figref{fig:background}(A))).
Finding answers to such query requires imputation and prediction of multiple edges across two parallel paths, while also using logical set operations (\eg, intersection, union). \Figref{fig:background}(B) shows the query computation plan and to determine the entities that are the answers to such a complex multi-hop query, missing links typically need to be implicitly inferred \Figref{fig:background}(C).
Notice that both tasks are closely related to each other. Knowledge graph completion can be viewed as a special case of a multi-hop reasoning task when the query consists of a single relation (\eg, {\em France--CapitalOf--?}). Multi-hop reasoning is a strict generalization of knowledge graph completion with much broader applicability but with its own set of unique computational and scalability challenges. 

Currently there are no frameworks that support multi-hop reasoning on massive Knowledge graphs. For example, among many recent works on multi-hop reasoning~\citep{hamilton2018embedding,ren2020query2box,ren2020beta,sun2020faithful,guu2015traversing,das2017chains,chen2021fuzzy,liu2021neural,kotnis2020answering,zhang2021cone,choudhary2021probabilistic} the largest KG used has only 63K entities and 592K relations.
Moreover, while there are scalable frameworks for single-hop KG completion~\citep{zhu2019graphvite,lerer2019pytorch,zheng2020dgl,pgl,mohoney2021marius}, such frameworks cannot be directly used for multi-hop reasoning due to the more complex nature of the multi-hop reasoning task.

Scaling up embedding-based multi-hop KG reasoning methods is a critical need for many real-world AI applications and remains largely unexplored. Two significant challenges exist: (1) on the algorithmic side, given a massive KG (with hundreds of millions of entities), it is no longer feasible to materialize training instances, and training data needs to be efficiently sampled on the fly with a high throughput to ensure GPUs are fully utilized. And (2), on the system side, recent single-hop large-scale KG embedding frameworks are based on graph-partitioning~\citep{zhu2019graphvite,lerer2019pytorch,zheng2020dgl,mohoney2021marius} which is problematic for multi-hop reasoning. Multi-hop reasoning requires traversing multiple relations in the graph, which will often span across multiple partitions.

\begin{figure*}[t]
\centering
\includegraphics[width=0.88\textwidth]{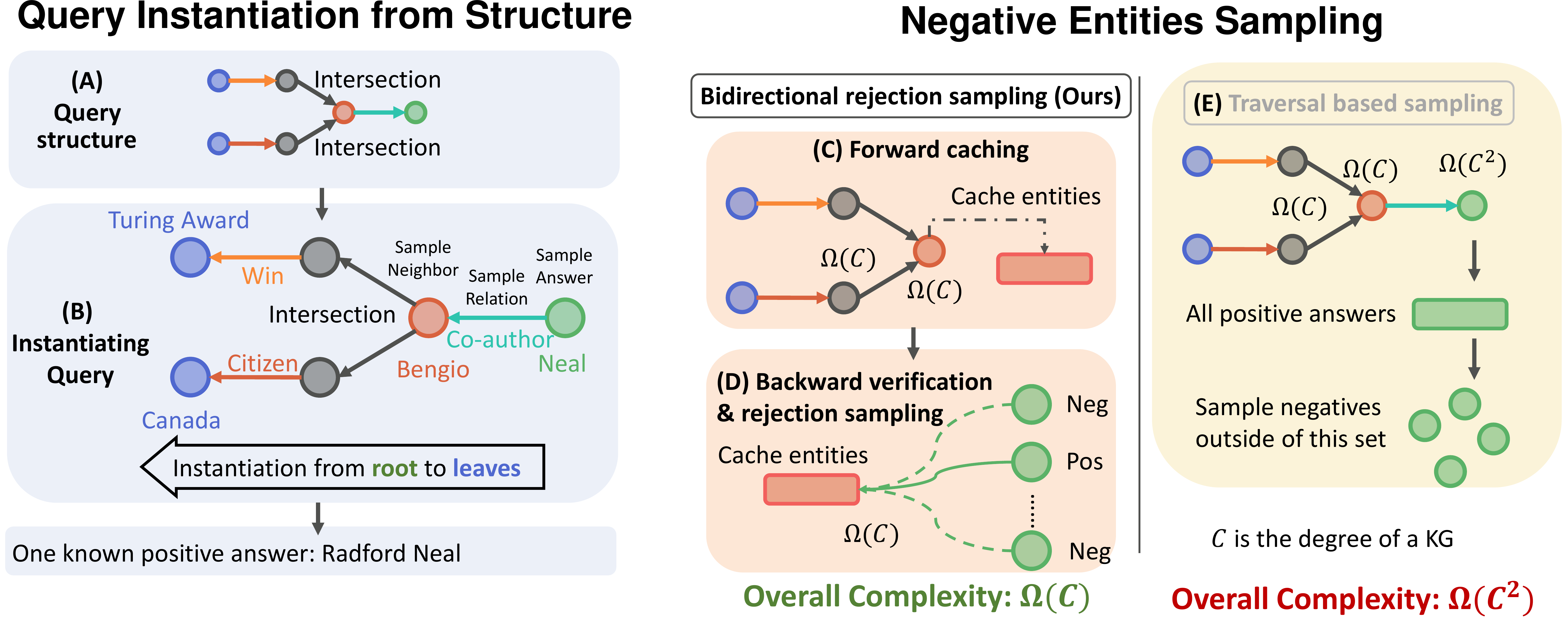}	
\vspace{-3mm}
\caption{Overview of the sampling process for query and positive/negative answers. Our system instantiates queries using query structures from root to leaves. The entity in the root naturally becomes positive answer to the instantiated query. For negative entities, we propose bidirectional rejection sampling, which has a square root computation complexity compared to the traversal-based method.
\label{fig:query} }
\vspace{-3mm}
\end{figure*}

To combat these challenges, we propose \emph{\modellong~(\modelshort)}, the \emph{first general framework} for single- and multi-hop reasoning on massive KGs. %
\modelshort performs algorithm-system co-optimization for scalability. \textbf{On the algorithmic side}, the key is to efficiently generate training examples online. To generate a training example with a set of positive and negative entities, we first instantiate a query on a given KG (\Figref{fig:query}(B)) from a set of query logical structures (\Figref{fig:query}(A)). The root of the instantiated query represents a known positive (answer) entity.
To obtain a set of negative entities (non-answers), 
n\"aive execution of the query computation plan (\Figref{fig:background}(B))
using KG traversal (\Figref{fig:background}(C)) to identify positive/negative entities has exponential complexity with respect to the number of hops of the query.
Therefore, we propose a bidirectional rejection sampling approach to efficiently obtain high-quality negative entities for the instantiated queries.  
The key insight of the training data sampler is to identify the \emph{optimal node cut} (red node in~\Figref{fig:query}(C)) of the computation plan via dynamic programming, then performing forward KG traversal (\Figref{fig:query}(C)) as well as backward verification (\Figref{fig:query}(D)) simultaneously, hence bidirectional rejection sampling. 
The nodes in the optimal cut cache the intermediate results from the forward KG traversal; for backward verification, we propose positive and negative candidate entities, traverse backward to the optimal cut and perform rejection sampling based on the overlap of the forward and backward sets. This reduces the worst case complexity by a square root, which makes it feasible to generate a training query, a positive answer entity and negative non-answer entities on the fly. 

\textbf{On the system side}, \modelshort operates on the full KG directly in a shared memory environment with multiple GPUs, while storing embedding parameters in the CPU memory to overcome the limited GPU memory. This design choice bypasses the potential drawbacks of graph partitioning for multi-hop reasoning in current KG embedding systems, but also brings efficiency challenges. We design an asynchronous scheduler to maximize the throughput of GPU computation, via overlapping sampling, asynchronous embedding read/write, neural network feed-forward, and optimizer updates, as depicted in Figure~\ref{fig:pipeline}. 
As a result, we obtain an efficient implementation that can achieve near linear speed-up with respect to the number of GPUs.

We demonstrate the scalability of \modelshort on three multi-hop reasoning algorithms (GQE~\citep{hamilton2018embedding}, Q2B~\citep{ren2020query2box}, BetaE~\citep{ren2020beta}), six single-hop KG completion approaches (TransE~\citep{bordes2013translating}, RotatE~\cite{sun2019rotate}, DistMult~\cite{yang2014embedding}, ComplEx~\cite{trouillon2016complex}, Q2B~\citep{ren2020query2box} and BetaE~\citep{ren2020beta}) on six different benchmark KGs. The largest is the Freebase KG~\cite{bollacker2008freebase,zheng2020dgl} that contains 86M nodes and 338M edges, which is about 1,500 times larger than the largest KG previously considered for multi-hop reasoning. The new sampling technique improves the worst case runtime of enumerative search by 4 orders of magnitude. On small KGs, \modelshort{} runs 2.2 times faster with 30.6\% less GPU memory usage compared to the existing (non-scalable) multi-hop reasoning framework~\citep{kgreasoning}. 
\modelshort also supports single-hop reasoning. Here \modelshort achieves comparable or even better efficiency with state-of-the-art frameworks (which do not support multi-hop reasoning) on both single-GPU and multi-GPU settings.

\modelshort can be easily deployed in a single-machine environment with a minimum requirement on the capacity of GPU memory. For example, it uses less than 2GB GPU memory when training a 400 dimensional embedding on the Freebase KG with 86M entities. \modelshort's throughput scales nearly linearly with the number of GPUs. 
\modelshort also provides an easy-to-use interface where implementing a new embedding model takes less than 50 lines of code. Together with the benchmarks, \modelshort provides a test bed for accelerating future research in multi-hop reasoning on KGs. \modelshort is open sourced at \url{https://github.com/google-research/smore}.


\section{Multi-hop reasoning on KG}\label{sec:background}
A knowledge graph (KG) $\Gcal=(\Vcal, \Ecal, \Rcal)$ consists of a set of nodes $\Vcal$, edges $\Ecal$ and relations $\Rcal$. Each edge $e\in\Ecal$ represents a triple $(v_h,r,v_t)$ where $r\in\Rcal$ and $v_h,v_t\in\Vcal$. We are interested in performing multi-hop reasoning on KGs. Multi-hop reasoning queries include relation traversals as well as several logical operations including conjunction ($\wedge$), disjunction ($\vee$), existential quantification ($\exists$) and negation ($\neg$). Here we consider first order logical queries in their disjunctive normal form~\citep{ren2020beta}. 
\begin{definition}[Logical queries~\citep{ren2020beta}]\label{def:query}
A first-order logical (FOL) query $q$ consists of a non-variable anchor entity set $\Vcal_q \subseteq \Vcal$, existentially quantified bound variables $V_1, \dots, V_k$ and a single target variable $V_?$ (answer). The disjunctive normal form of a query $q$ is defined as follows: 
\begin{align*}
\centering
	\textstyle
    q[V_{?}] = V_?\:.\:\exists V_1, \ldots, V_k : c_1 \vee c_2 \vee ... \vee c_n
\end{align*}
\begin{itemize}[noitemsep,topsep=0pt,parsep=0pt,partopsep=0pt, leftmargin=*]
    \item Each $c$ represents a conjunction of one or more literals $e$. $c_i = e_{i1} \wedge e_{i2} \wedge \dots \wedge e_{im}$.
    \item Each $e$ represents an atomic formula or its negation. $e_{ij} = r(v_a, V)$ or $\neg\:r(v_a, V)$ or $r(V', V)$ or $\neg\:r(V', V)$, where $v_a \in \Vcal_q$, $V \in \{V_?,V_1,\ldots,V_k\}$, $V' \in \{V_1,\ldots,V_k\}$, $V\neq V'$, $r \in \Rcal$.
\end{itemize}
\end{definition}

\paragraph{Query computation plan.} 
A query computation plan (\Figref{fig:background}(B)) provides a plan for executing the query. The computation plan consists of nodes $\Vcal_q \cup \{V_1, \dots, V_k, V_?\}$, where each node corresponds to \emph{a set of entities} on the KG. The edges in the computation plan represent a logical/relational transformation of this set, including relation projection, intersection, union and complement/negation. We adopt the same definition of computation plan as in \cite{ren2020beta} (details in \appref{app:computegraph}).

\paragraph{Traversing the KG using the computation plan to find answers.}
Conceptually, (assuming no noise, no missing relations in the KG) a logical query can be answered by traversing the edges of the KG. For a valid query, the computation plan is a tree, where the anchor entity set, $\Vcal_q$, are the leaves and the target variable $V_?$ is the single root, representing the set of answer entities (\Figref{fig:background}(C)). 
Following the computation plan, we start with the anchor entities, traverse the KG and execute logical operators towards the root node. The answers $\Acal_q^\Gcal$ to the query $q$ are stored in the root node after the KG traversal. 
Note that this conceptual traversal would have exponential computational complexity with respect to the number of hops and also cannot handle noisy or missing relations in the KG, which are both very common in real-world KGs (\Figref{fig:background}(C)). 

\paragraph{Embedding-based ``traversal'' of the KG.} Embedding-based reasoning methods avoid explicit KG traversal. Instead, they start with the embeddings of anchored entities, and then apply a sequence of neural logical operators according to the query computation plan. This way we obtain the embedding of the query where each embedding-based logical operator (\eg, negation) takes the current input embedding and transforms it into a new output embedding. Such operators are then combined according to the query structure. The answers to the query $q$ are then entities $v$ that are embedded close to the final query embedding. The distance is measured by a pre-defined function $\texttt{Dist}(f_\theta(q),f_\theta(v))$, where $f_\theta(q)$ and $f_\theta(v)$ represents the query and entity embedding respectively. Note the distance function $\texttt{Dist}(\cdot,\cdot)$ is tailored to different embedding space and model design $f_\theta$, as specified in~\tabref{tab:kg-model} (\appref{appendix:models} for details).

\begin{table*}[t]
\centering
\caption{Three multi-hop KG reasoning models supported by \modelshort. 
\label{tab:kg-model}}
\vspace{-4mm}
\resizebox{1.9\columnwidth}{!}{%
\begin{tabular}{cccccc}
	\toprule
Model & Embedding Space & Relation Projection & Intersection & Negation & Distance \\
\hline
GQE~\citep{hamilton2018embedding} & $\Em{q}\in\RR^d$, $\Em{v}\in\RR^d$ & $\Em{q}+\Em{r}$ & $\texttt{DeepSet}(\{\Em{q_i}\})$ & - & $\|\Em{q}-\Em{v}\|$ \\\hline
\multirow{2}{*}{Q2B~\citep{ren2020query2box}} &  \multirow{2}{*}{$\Em{q}\in\RR^{2d}$, $\Em{v}\in\RR^d$}& \multirow{2}{*}{$\Em{q}+\Em{r}$} & $\Ce{q}=\sum_i \mathbf{a}_i\odot\Ce{q_i}$ & \multirow{2}{*}{-} & \multirow{2}{*}{$\text{dist}_\text{out}+\alpha\text{dist}_\text{in}$}\\
&&&$\Of{q}=\text{min}(\{\Of{q_i}\})\odot\sigma(\texttt{DeepSet}(\{\Of{q_i}\}))$&&\\\hline

BetaE~\cite{ren2020beta} & $\Em{q}\in\RR^{d}$, $\Em{v}\in\RR^{d}$ & $\texttt{MLP}(\Em{q},\Em{r})$ & $\Em{q}=[(\sum w_i\alpha_i,\sum w_i \beta_i)]$ & $\frac{1}{\Em{q}}$ & $\texttt{KL}(\text{Beta}(\Em{v});\text{Beta}(\Em{q}))$ \\
	\bottomrule
\end{tabular}
}
\end{table*}


\begin{table}[t]
\centering
\vspace{-2mm}
\caption{Six single-hop models supported by \modelshort. 
\label{tab:single-kg-model}}
\vspace{-3mm}
\resizebox{0.9\columnwidth}{!}{%
\begin{tabular}{ccc}
	\toprule
Model & Embedding Space & Distance \\
\hline
TransE~\citep{bordes2013translating} & $\Em{h},\Em{t}\in\RR^d$, $\Em{r}\in\RR^d$ & $\|\Em{h}+\Em{r}-\Em{t}\|$ \\
\hline
RotatE~\cite{sun2019rotate} & $\Em{h},\Em{t}\in\CC^d$,$\Em{r}\in\CC^{d}$ & $\|\Em{h}\circ\Em{r}-\Em{t}\|$ \\\hline
DistMult~\cite{yang2014embedding} & $\Em{h},\Em{t}\in\RR^d$,$\Em{r}\in\RR^{d}$ & $-<\Em{h}\circ\Em{r},\Em{t}>$ \\\hline
ComplEx~\cite{trouillon2016complex} & $\Em{h},\Em{t}\in\CC^{d}$, $\Em{r}\in\CC^{d}$ & $-\text{Re}(<\Em{h}\circ\Em{r},\overline{\Em{t}}>)$ \\\hline
Q2B~\cite{ren2020query2box} & $\Em{h},\Em{t}\in\RR^{d}$, $\Em{r}\in\RR^{2d}$ & $\text{dist}_\text{out}+\alpha\text{dist}_\text{in}$ \\\hline
BetaE~\cite{ren2020beta} & $\Em{h},\Em{t}\in\RR^{d}$, $\Em{r}\in\RR^{d}$ & $\texttt{KL}(\text{Beta}(\Em{t});\text{Beta}(\texttt{MLP}(\Em{h},\Em{r})))$ \\

	\bottomrule
\end{tabular}
}%
\end{table}

\textbf{Contrastive learning for KG embeddings.}
During training, we are given a data sampler $\Dcal$, each sample in $\Dcal$ is a tuple $(q,\Acal_q^\Gcal,\Ncal_q^\Gcal)$, which represents a query $q$, its answer entities $\Acal_q^\Gcal\subseteq \Vcal$ and the negative samples $\Ncal_q^\Gcal\subseteq\overline{\Acal_q^\Gcal}$. The contrastive loss~\eqnref{eq:lossfunc} is designed to minimize the distance between the query embedding and its answers $\texttt{Dist}(f_\theta(q), f_\theta(v)), v\in\Acal_q^\Gcal$ while maximizing the distance between the query emebdding and the negative samples $\texttt{Dist}(f_\theta(q), f_\theta(v')), v'\in\Ncal_q^\Gcal$,
\begin{align} 
    &\Lcal\rbr{\theta} = - \frac{1}{\abr{\Acal}}\sum_{v\in \Acal_q^\Gcal}\log \sigma \left(\gamma - \texttt{Dist}(f_\theta(q),f_\theta(v)) \right) \nonumber \\
    &- \frac{1}{\abr{\Ncal}}\sum_{v'\in\Ncal_q^\Gcal} \log \sigma \left(\texttt{Dist}(f_\theta(q),f_\theta(v'))-\gamma \right),\label{eq:lossfunc}
\end{align}
where $\gamma$ is a hyperparameter that defines the margin and $\sigma$ is the sigmoid function. 

We emphasize that due to the multi-hop structure in reasoning, identifying/computing $\Acal_q^\Gcal$ and $\Ncal_q^\Gcal$ involves complex first-order logical operations, which are significantly more expensive than sampling in classical (single link) KG completion tasks, and thus is the bottleneck for scaling-up. 
To resolve this, we propose \modellong~(\modelshort) to scale up single- and multi-hop KG reasoning methods (Tables~\ref{tab:kg-model} and \ref{tab:single-kg-model}), with an efficient sampling algorithm and parallel training, for a given constrastive loss. In then next section we first discuss our efficient training data sampling algorithm.

\begin{figure*}[t]
\centering
\includegraphics[width=0.9\textwidth]{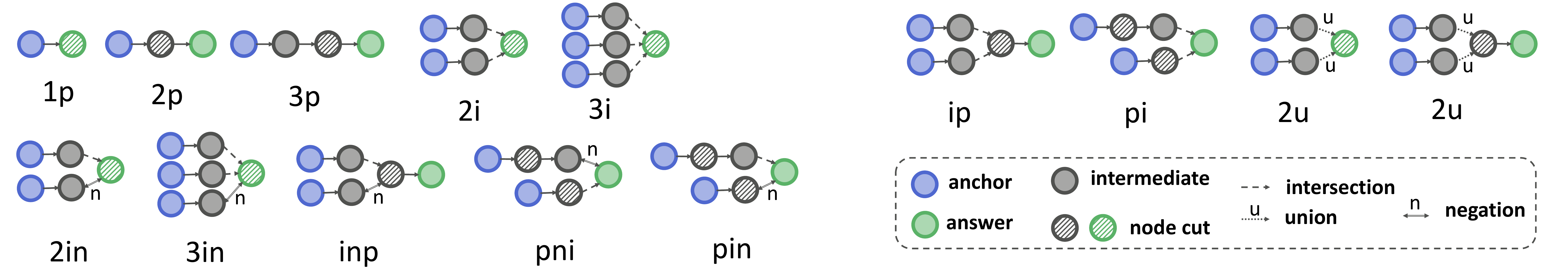}
\vspace{-2mm}
\caption{Different query structures and their optimal node cuts (shaded nodes) used by our bidirectional rejection sampling. 
\label{fig:query_tree}}
\vspace{-7mm}
\end{figure*}

\section{Efficient training data sampling}\label{sec:sampler}
Unlike in link prediction, where sampling the training data, {\em head--relation--tail} can be quickly performed via dictionary look up~\citep{zhu2019graphvite, zheng2020dgl}, sampling training data for multi-hop reasoning is much more complicated. It involves generating queries $q$ by instantiating query structures, performing KG traversal to find answers $\Acal_q^{\Gcal}$ as well as negative answers $\Ncal_q^{\Gcal}$, which is computationally expensive. We propose an efficient way to sample training data for contrastive learning for multi-hop reasoning. During inference, there are usually a pre-generated test set or user can input the test query of interest. This section focuses on efficient sampling during training.

\subsection{Instantiating query structure}
\label{sec:inst_q}

A query logical structure (\Figref{fig:query_tree}) specifies the backbone of a query $q$, including the types of operation (intersection, relation projection, negation, union) and its structure. It can be seen as an abstraction of the query computation plan, where anchor nodes and relation types are not grounded. Instantiating a query structure requires specifying a relation $r\in\Rcal$ for each edge in the structure, as well as the anchor entities $\Vcal_q$ (blue nodes in \Figref{fig:query_tree}). 

A na\"ive way to instantiate a query (construct a concrete query given its logical structure) is to first ground the anchor entities by randomly sampling entities in the KG, and then randomly select relations $r \in \Rcal$ for all the relation projection edges. However, in most cases such randomly generated queries have no answers in the KG as sampled entities may not even have relations of predetermined types, and intersections of random entities will almost always be empty. This means such samples have to be rejected and the sampling process has to start all over again. Such a na\"ive method leads to huge computation cost.

Instead, we instantiate a query structure via {\em reverse directional sampling}, \ie, we first ground the root node (\ie, the answer node) and then proceed towards the anchors. This is the reverse process of KG traversal for answers (\Secref{sec:background}). The main benefit of reverse sampling is that it can always instantiate a given query structure (\appref{app:reverse_sample}). Reverse directional sampling uses depth-first search (DFS) over the query structure from the root (answer) to the leaves (anchor entities). During the DFS, each node on the query structure is grounded to an entity on the KG and an edge to a relation on the KG associated with the previously grounded entity. An example of the process is shown in~\Figref{fig:query}(B). 

Reverse sampling procedure can always obtain valid queries (with non-empty answer set). Another advantage of the above sampling process is that, the overall complexity is $O(C|q|)$, same as the complexity of DFS, where $|q|$ indicates the maximum depth of a path (from root to leaves) in the query structure, and $C$ is the maximum degree of entities in the KG. The sampling process returns the instantiated query $q$, the anchor entities $\Vcal_q$, and a single positive answer $a_q \in \Acal_q^{\Gcal}$ (the instantiated entity at root).

\subsection{Negative sampling}
\label{sec:bidir_sampling}

After the instantiating (node/edge grounding) the query structure, we obtain the tuple $(q, \Vcal_q, \cbr{a_q})$ as a positive sample while we still need $\Ncal_q^{\Gcal}$ to optimize~\eqnref{eq:lossfunc}. We find that a single answer entity is sufficient in each step of stochastic training, while typically we need $k = |\Ncal_q^{\Gcal}|$ in~\eqnref{eq:lossfunc} to be thousands for negative samples in the contrastive learning objective. Next we explain how to efficiently obtain a set of negative entities $\Ncal_q^{\Gcal}$ (\ie, non-answers).

A na\"ive approach would be to sample negative entities (non-answers) at random from the KG, independent of the query $q$. However, notice that a valid query may have many answers entities in the order of $O(C^{|q|})$. Such an approach would mean that many of the sampled negatives are actually answers to the query, which would lead to noisy training data that would confuse the model.
An alternative would be to execute the query $q$ and perform KG traversal to obtain all the answers $\Acal_q^\Gcal$ (as presented in \Secref{sec:background}). We could then obtain negative samples $\Ncal_q^\Gcal$ by simply sampling from $\Vcal \backslash\Acal_q^\Gcal$. Although it is possible to re-order the relation projection operations to get better scheduling, in the worst case, $|\Acal_q^{\Gcal}|$ is still in the order of $O(C^{|q|})$, regardless of the query scheduling. Thus, such exhaustive traversal is prohibitive for negative sampling on large KGs.

\textbf{Our solution: Bidirectional rejection sampling.}
Since $\Ncal_q^{\Gcal}$ does not have to contain all the non-answer entities during stochastic training, we propose to exploit rejection sampling to locate a subset of negative entities efficiently. That is to say, starting with a random proposal $v \in \Vcal$, we only need to check whether $v \in \Acal_q^{\Gcal}$, rather than to enumerate the entire $\Acal_q^{\Gcal}$. Inspired by bidirectional search, our key insight is to obtain a node cut (formally defined in Def. \ref{def:nodecut}) on the query computation plan, \ie, a subset of nodes that cut all the paths between each leaf node and the root node. Then we perform bidirectional search. We first start the traversal from the leaves (anchors) to the node cut and cache the entities we obtained in traversal, which we term \emph{forward caching} (\Figref{fig:query}(C)). We then sample negative entities, traverse from the root to the node cut and verify whether they are true negatives by checking the overlap of the cached entities and the traversed set. We term the second process \emph{backward verification} (\Figref{fig:query}(D)).

\begin{definition}[Node cut]\label{def:nodecut}
	A node cut $c_q$ of a query $q$ is a set of nodes in the computation plan, such that every path between anchor node (leaf) and answer node (root) contains exactly one node in $c_q$. By definition a node cut is minimal, meaning that no subset of $c_q$ can be a node cut.
\end{definition}

We illustrate this idea in~\Figref{fig:query}. Given a two-hop query ``\textit{Who co-authored papers with Canadian Turing Award winners?}'', we set the node after the intersection operation (red node) as the single node in the node cut. Then we can obtain the set of ``Canadian Turing Award winners'' via \textit{forward traversal} and cache the intermediate results (\ie, \texttt{Bengio}). 
Overall this process requires $O(C)$ computation/memory cost, where $C$ is the degree of the KG. 
Then given a candidate negative entity $v$, one can spend $O(C)$ cost to verify whether the set of co-authors of $v$ overlaps with the cached entities in the node cut.
In our implementation (\secref{sec:more_opt} for more details) we propose constant number of candidate negative entities, thus the overall computation cost would be $O(C)$, which is a reduction of square root from $O(C^2)$ using exhaustive traversal. 

Next, we calculate the computation cost for any given node cut $c_q$, and then we propose an efficient algorithm to find the optimal node cut, \ie, one with the lowest cost in bidirectional search.

Given a reasoning path $P_{(v_a, V_?)} = [v_0 = v_a, v_1, \ldots, v_t = V_?]$ in the query computation plan that starts from an anchor node (leaf) $v_a \in \Vcal_q$ and ends at the answer node (root) $V_?$, for a node cut $c_q$, by definition there exists a unique node $v_i \in c_q \cap P_{(v_a, V_?)}$. Then the worst-case computation/memory cost for negative sampling for reasoning path $P_{(v_a, V_?)}$ can be estimated as $\texttt{cost}(c_q, P_{(v_a, V_?)}) = \max\cbr{C^{i}, C^{t-i}}$, \ie, the maximum cost of forward traversal or backward verification. The optimal scheduling is recast as 
\begin{eqnarray}
	\min_{c_q} \quad & \max_{v_a \in \Vcal_q} \texttt{cost}(c_q, P_{(v_a, V_?)}) \nonumber \\
	\quad & \st \,\, c_q \text{ is a node cut of } q. \label{eq:optimal_sheduling}
\end{eqnarray}
As the computation plan is a tree, we propose to solve the above optimization problem with dynamic programming (DP). We define three functions $u(v)$, $s(v)$, $o(v)$ representing the number of relation projections from $v$ to the root $V_?$, the maximum length of path from $v$ to any anchor, and the optimal cost of resolving all the reasoning paths that include $v$ in the best plan. We can derive the dependency of the three functions recursively, solve the dynamic program in a linear time w.r.t. $|q|$, and construct the node cut using the function $o(\cdot)$. We provide the details in \appref{app:cut_dp}. Example query structures and their corresponding optimal node cuts are shown in \Figref{fig:query_tree}\footnote{Although query structures considered in current literature are small enough to find the optimal cut with brute force, our DP significantly improves efficiency when query structures are large.}.


\section{Efficient training system}
\label{sec:system}

\begin{figure}[t!]
\centering
\includegraphics[width=0.4\textwidth]{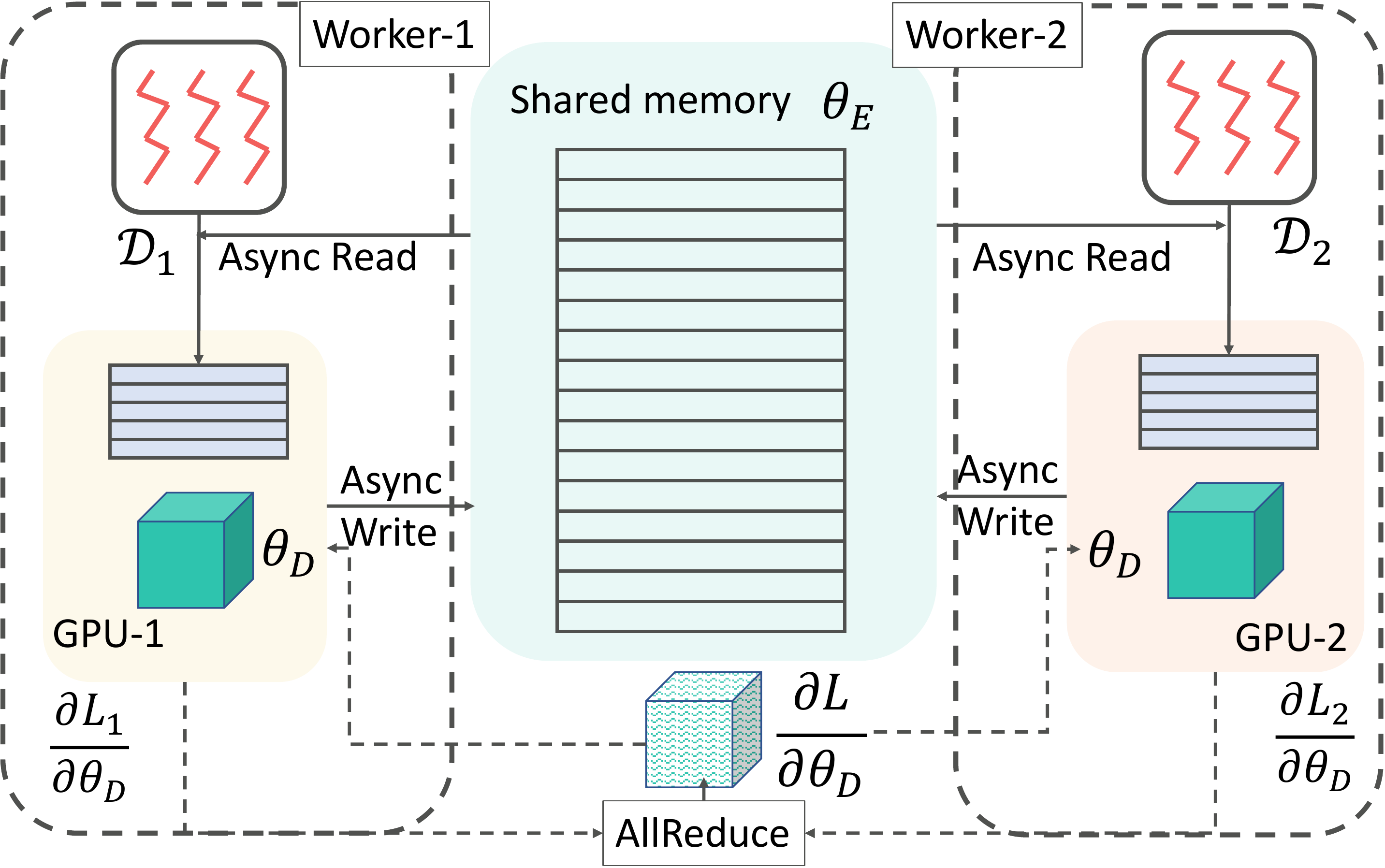}
\vspace{-2mm}
\caption{Overall training paradigm of \modelshort.  \label{fig:training_paradigm}}
\end{figure}

\begin{figure*}[t]
\centering
    \includegraphics[width=0.9\textwidth]{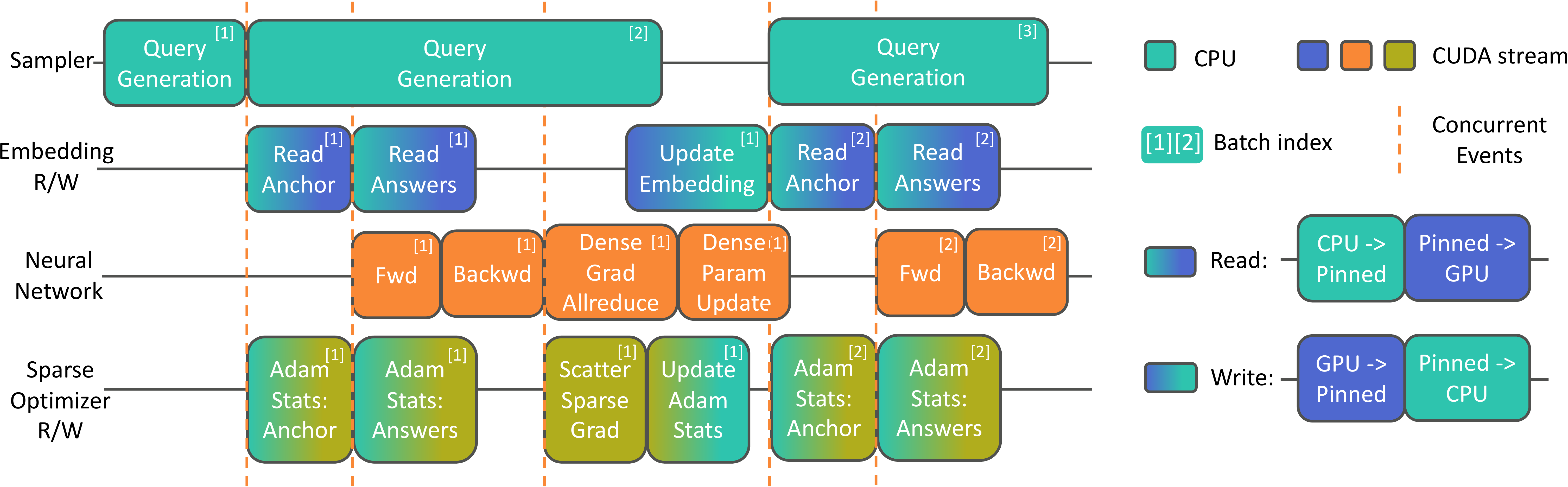}
    \caption{\modelshort pipeline of a single worker process. \label{fig:pipeline}}
    \vspace{-3mm}
\end{figure*}

\begin{table*}[t]
\centering
\caption{KG statistics with number of entities, relations, training, validation and test edges.\label{tab:meta_kg} }
\small
\resizebox{1.8\columnwidth}{!}{%
\begin{tabular}{|l|c|c|c|c|c|c|}
\hline
\textbf{Dataset} & \textbf{Entities} & \textbf{Relations} & \textbf{Training Edges} & \textbf{Validation Edges} & \textbf{Test Edges} & \textbf{Total Edges}\\
\hline
FB15k & 14,951 & 1,345 & 483,142 & 50,000 & 59,071 & 592,213\\
\hline
FB15k-237 & 14,505 & 237 & 272,115 & 17,526 & 20,438 & 310,079\\
\hline
NELL995 & 63,361 & 200 & 114,213 & 14,324 & 14,267 & 142,804\\
\hline
FB400k & 409,829 & 918 & 1,075,837 & 537,917 & 537,917 & 2,151,671 \\
\hline
ogbl-wikikg2 & 2,500,604 & 535 & 16,109,182 & 429,456 & 598,543 & 17,137,181 \\
\hline
Freebase & 86,054,151 & 14,824 & 304,727,650 & 16,929,318 & 16,929,308 & 338,586,276 \\
\hline
\end{tabular}
}%
\end{table*}

\modelshort is built for a shared memory environment with multi-cores and multi-GPUs. It combines the usage of CPU and GPU, where the dense matrix computations are deployed on GPUs, and the sampling operations are on CPUs.

\subsection{Distributed training paradigm}
\label{sec:dist_train}

Here we give a high-level introduction to the distributed training. 
Most of the KG embedding methods would maintain an embedding matrix $\theta_E \in \RR^{|\Vcal| \times d}$, where $d$ is the embedding dimension which can typically be 512 or larger. For a large KG with more than millions of entities, the embeddings $\theta_E$ cannot be stored in GPUs, since most GPUs would have 16GB or lower memory. 
Thus similar as recent works~\citep{zheng2020dgl}, we put the embedding matrix on shared CPU memory, while putting a copy of other parameters $\theta_D = \theta \setminus \theta_E$, \eg, neural logical operators, in each individual GPU.

We launch one worker process per GPU device. For simplicity, we use subscript $w$ as an index of a worker. Worker $w$ gets the shared access to $\theta_E$ and local GPU copy of dense parameters $\theta_D$. Each worker repeats the following stages as shown in \Figref{fig:training_paradigm} until the training stops:
\begin{enumerate}[noitemsep,topsep=0pt,parsep=1pt,partopsep=0pt, leftmargin=*]
	\item Collect a mini-batch of training samples $\cbr{D_i}_w$ from $\Dcal_w$, which is the sampler.
	\item Load relevant entity embeddings from CPU to GPU;
	\item Compute gradients locally, and perform gradient \texttt{AllReduce} using $\frac{\partial \Lcal_w}{\partial \theta_D}$. Update local copy $\theta_D$.
	\item Update shared $\theta_E$ asynchronously with $\frac{\partial \Lcal_w}{\partial \theta_E}$. 
\end{enumerate}
In the shared memory with multi-GPU scenario, the heavy CPU/GPU memory read/write with $\theta_E$ is necessary for every round of stochastic gradient update. This significantly lowers the FLOPS on GPU devices if we execute the above stages in a serialized way. So instead, we design the asynchronous pipeline that will be covered in \Secref{sec:async_design}.

The different storage location of parameters also brings different read/update mechanisms. For the embedding parameters $\theta_E$, as only a tiny portion will be accessed during each iteration in stochastic training, the asynchronous update on the shared CPU memory would still result in a convergent behavior~\citep{niu2011hogwild}. Unlike link prediction models, most multi-hop reasoning models are additionally equipped with dense neural logical operators, used in all batches and iterations.
To minimize the loss of performance of multi-GPU training of these dense parameters, we choose to synchronously update $\theta_D$ with \texttt{AllReduce} operations that are available in multi-GPU environment with the NVIDIA Collective Communication Library (NCCL).

\subsection{Asynchronous design}
\label{sec:async_design}

In this section, we present the asynchronous mechanism for pipelining the stages in each stochastic gradient update. The stages can be virtually categorized into four kinds of \textit{meta-threads}, where each kind of meta-thread may consist of multiple CPU threads or CUDA streams. These meta-threads run concurrently, with possible synchronization events for pending resources, as illustrated in~\Figref{fig:pipeline}.
Below we will elaborate each type of meta-thread individually.

\textbf{Multi-thread sampler.}
Each worker $w$ maintains one sampler $\Dcal_w$ that has access to the shared KG. The sampler contains a thread pool for sampling queries and the corresponding positive/negative answers in parallel. 
The data sampler works concurrently with the other meta-threads. The pre-fetching mechanism will obtain samples for the next mini-batch while training happens using current batch on other threads. So if the sampler is efficient enough, then the runtime can almost be ignored.

\textbf{Sparse embedding read/write.}
For the embedding matrix $\theta_E$, we will create a single background thread with a CUDA stream for embedding read and write. Specifically, when loading the embedding of some entities into GPU, the background thread first loads that into a pinned memory area, then the CUDA asynchronous stream will perform pinned memory to GPU memory copy \cite{fey2021gnnautoscale}. This read operator is non-blocking, and will not be synchronized until the CUDA operator in the main CUDA stream asks for it. The write operation works similarly but in the reverse direction. It is also possible to have multiple background threads. Right part of \Figref{fig:pipeline} illustrates the idea. 

\textbf{Dense computation.}
The feed-forward of model $f_{\theta}$ starts when training data $(q, \Vcal_q, \Acal_q^{\Gcal}, \Ncal_q^{\Gcal})$ is ready and the embedding of the anchor entities $\Vcal_q$ is fetched into GPU. The embeddings of $\Acal_q^{\Gcal}$ and $\Ncal_q^{\Gcal}$ can be fetched as late as when we compute the loss function, in order to overlap the computation and memory copy. After obtaining the local gradients $\frac{\partial \Lcal_w}{\partial \theta_E}$ and $\frac{\partial \Lcal_w}{\partial \theta_D}$, the asynchronous update for $\theta_E$ will be invoked first without blocking,
and at the same time the \texttt{AllReduce} operation will start, followed by the dense parameter update of $\theta_D$ on the GPU.

\textbf{Sparse optimizer with asynchronous read/write.}
Different from $\theta_D$, only a small set of rows of $\theta_E$ will be involved in each stochastic update. So 
we only keep track of $\theta_E^{\Vcal_q}$, $\theta_E^{\Ncal_q^{\Gcal}}$, $\theta_E^{\Acal_q^{\Gcal}}$ and their gradients, \ie, the embeddings that are relevant to positive/negative and anchor entities. Once the back-propagation is finished, we will scatter $\frac{\partial \Lcal_w}{\partial \theta_E^{\Vcal_q}}$, $\frac{\partial \Lcal_w}{\partial \theta_E^{\Ncal_q^{\Gcal}}}$ and $\frac{\partial \Lcal_w}{\partial \theta_E^{\Acal_q^{\Gcal}}}$ into a single continuous memory, due to the potential overlap among the sets $\Vcal_q$, $\Ncal_q^{\Gcal}$ and $\Acal_q^{\Gcal}$.
We use Adam~\citep{kingma2014adam} for \modelshort, thus we need to additionally keep the first and second order moments of gradients in CPU. They are treated in the same way as $\theta_E$, and thus will have the same asynchronous read/write behavior as discussed above.
Every time a set of embeddings is retrieved from $\theta_E$, the optimizer will also start to pre-fetch the corresponding first/second order moments in a different background thread (``Adam Stats'' in \Figref{fig:pipeline}).


\begin{table*}[t]
\centering
\caption{Performance of \modelshort compared to KGReasoning package~\citep{kgreasoning} on small KGs. Under the same model configurations and hyperparameters, \modelshort achieves similar MRR (25.67 vs. 25.65) but with significant speed-up (2.2$\times$ faster training/throughput) and GPU memory saving (-30.6\%).
\label{tab:calibrate} }
\vspace{-3mm}
\resizebox{0.8\textwidth}{!}{%
\begin{tabular}{c|c|cc|cc|cc}
\toprule
\multirow{2}{*}{Dataset} & \multirow{2}{*}{Model} & \multicolumn{2}{c}{MRR (\%)} & \multicolumn{2}{c}{Training queries (x512/Sec)} & \multicolumn{2}{c}{GPU Memory (MB)} \\
\cline{3-8}
& & KGReasoning & \modelshort{} & KGReasoning & \modelshort{} & KGReasoning & \modelshort{} \\
\hline
\multirow{3}{*}{FB15k} & BetaE & 41.6 & 40.39 & 12.93 & 45.66 & 4022 & 1942 \\
& Q2B & 38.0 & 41.54 & 53.43 & 104.20 & 2146 & 1616 \\
& GQE & 28.0 & 30.60 & 55.67 & 118.05 & 2126 & 1778 \\
\hline
\multirow{3}{*}{FB15k-237} & BetaE & 20.9 & 19.67 & 12.92 & 45.46 & 4010 & 1876 \\
& Q2B & 20.1 & 20.42 & 55.82 & 107.05 & 2138 & 1582 \\
& GQE & 16.3 & 15.68 & 61.89 & 120.17 & 2116 & 1738 \\
\hline
\multirow{3}{*}{NELL995} & BetaE & 24.6 & 23.17 & 10.24 & 35.82 & 4852 & 3014 \\
& Q2B & 22.9 & 21.84 & 40.69 & 85.71 & 2406 & 2172 \\
& GQE & 18.6 & 17.53 & 41.32 & 68.81 & 3062 & 2922 \\
\hline
\multicolumn{2}{c}{Average} & 25.67 & 25.65 & 38.32 & 81.21 ({\color{darkpastelgreen} +2.2$\times$}) & 2986 & 2071 ({\color{darkpastelgreen} -30.6\%}) \\
\bottomrule
\end{tabular}
}%
\end{table*}


\subsection{Further optimization of \modelshort}
\label{sec:more_opt}

In addition to the above optimized system design for \modelshort, there are several other important optimization that further speeds up the training, which we highlight below.

\textbf{Sharing negative samples.}
Although with the asynchronous design we can overlap the embedding R/W with GPU computation, it is still important to keep the size of memory exchange small. Inspired by \citet{zheng2020dgl}, we share the negative answers among the queries in a mini-batch. Each mini-batch data is formatted as $\rbr{\Ncal, \cbr{(q_i, \Vcal_{q_i}, \Acal_{q_i}}_{i=1}^M, Mask}$, where $\Ncal \subset \Vcal$ are the shared negative answers for all queries. $Mask \in \cbr{0, 1}^{M \times |\Ncal|}$ is an indicator matrix. $Mask_{i, j}$ specifies whether the $j$-th entity in $\Ncal$ is a negative sample for $q_i$. 

This sharing trick also favors our bi-directional rejection sampler, as by nature it is the sample-and-check process, with the only adaptation that all the queries in the same mini-batch share the support of negative sampling proposal. 

\textbf{Customized CUDA distance kernel.}
Given $M$ queries and negative answer candidates $\Ncal$, the computation of $M \times |\Ncal|$ pairs of distances is model dependent. While generally distances like inner-product or L2 can be implemented with efficient matrix multiplication, geometries designed for multi-hop reasoning like box or beta distribution requires more complicated distance metrics like KL divergence. We provide a generic interface to parallelize the computation with customized CUDA kernel, which also enables the operation fusion to reduce GPU memory consumption.

\textbf{Scheduled structure sampling. }
Since we have the synchronized gradient update for $\theta_D$ at each step, it is important to balance the computation among worker processes. One aspect is the query tree structures sampled in each mini-batch. Some of the simple structures like 1-hop question can be computed very fast, while others like conjunctions with negations need not only multi-hop computation, but also involve dense neural network calculation. To balance the workload, we make the sampler $\Dcal_w$ of each worker $w$ to sample queries of the same structure in each mini-batch, by synchronizing the random seed at beginning. Note that we only make the query structure same, not the actual instantiated query.


\begin{figure*}[t]
\centering
\begin{tabular}{@{}c@{}c@{}c@{}}
\includegraphics[width=0.25\textwidth]{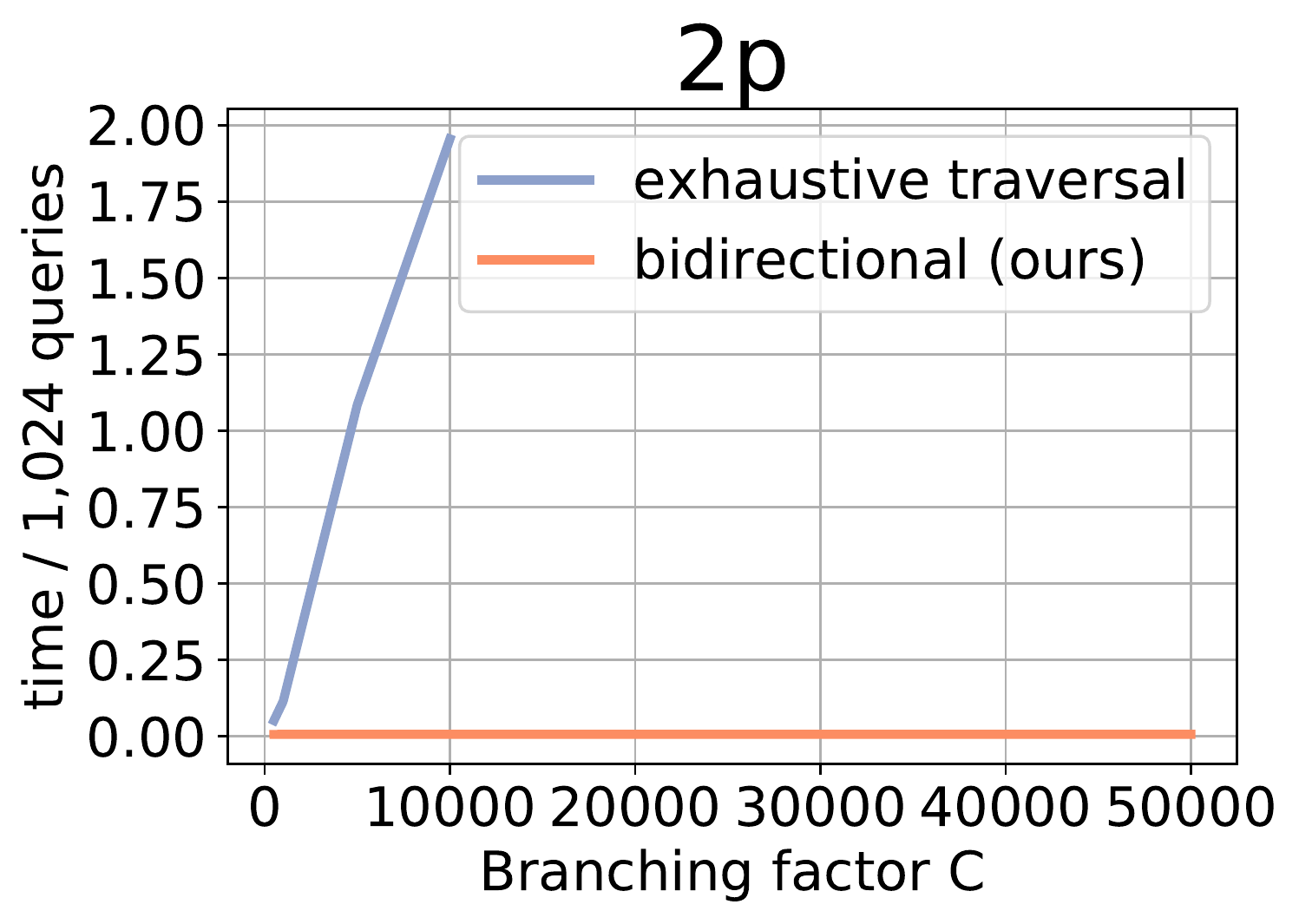} & 
\includegraphics[width=0.25\textwidth]{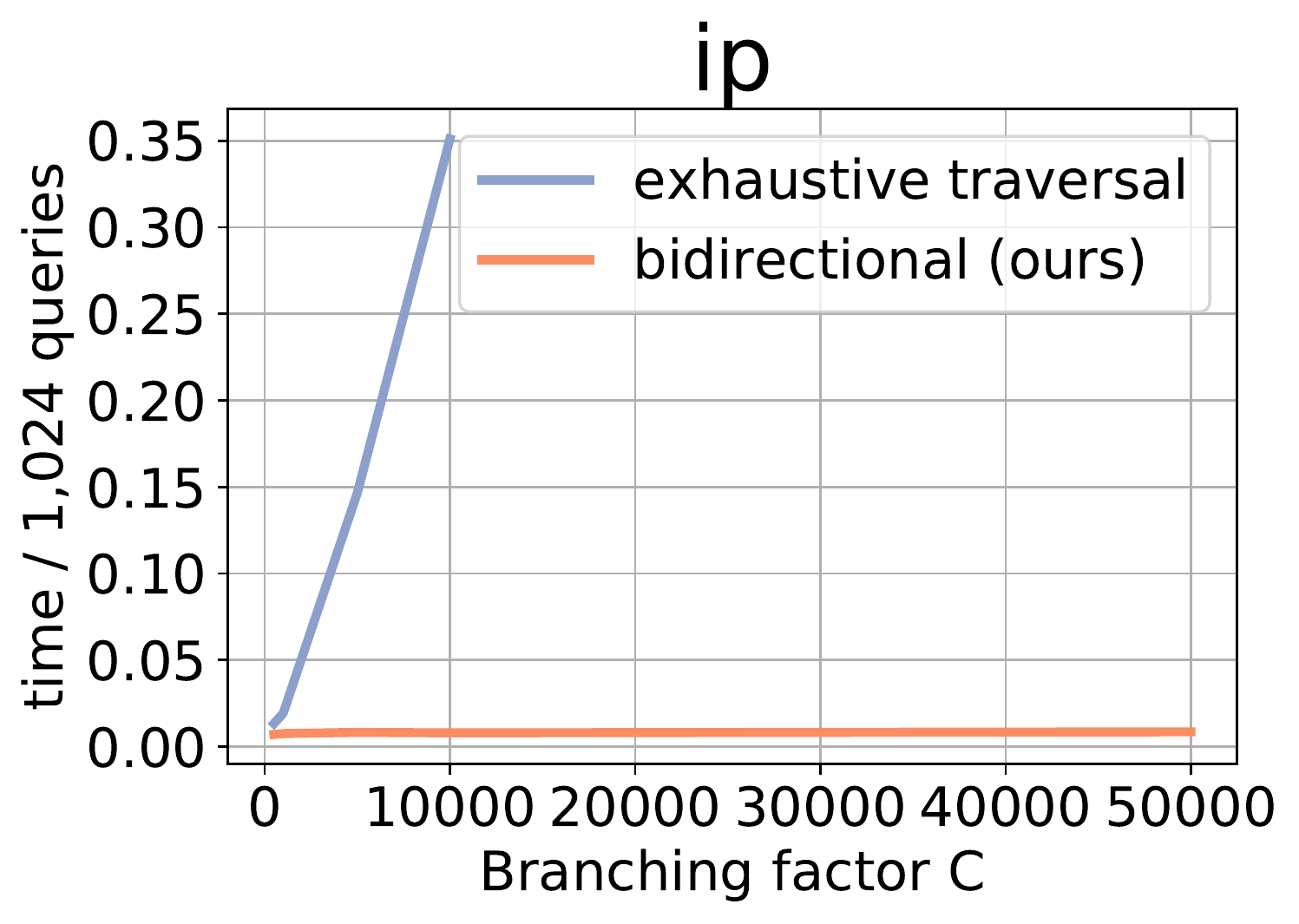} & 
\includegraphics[width=0.25\textwidth]{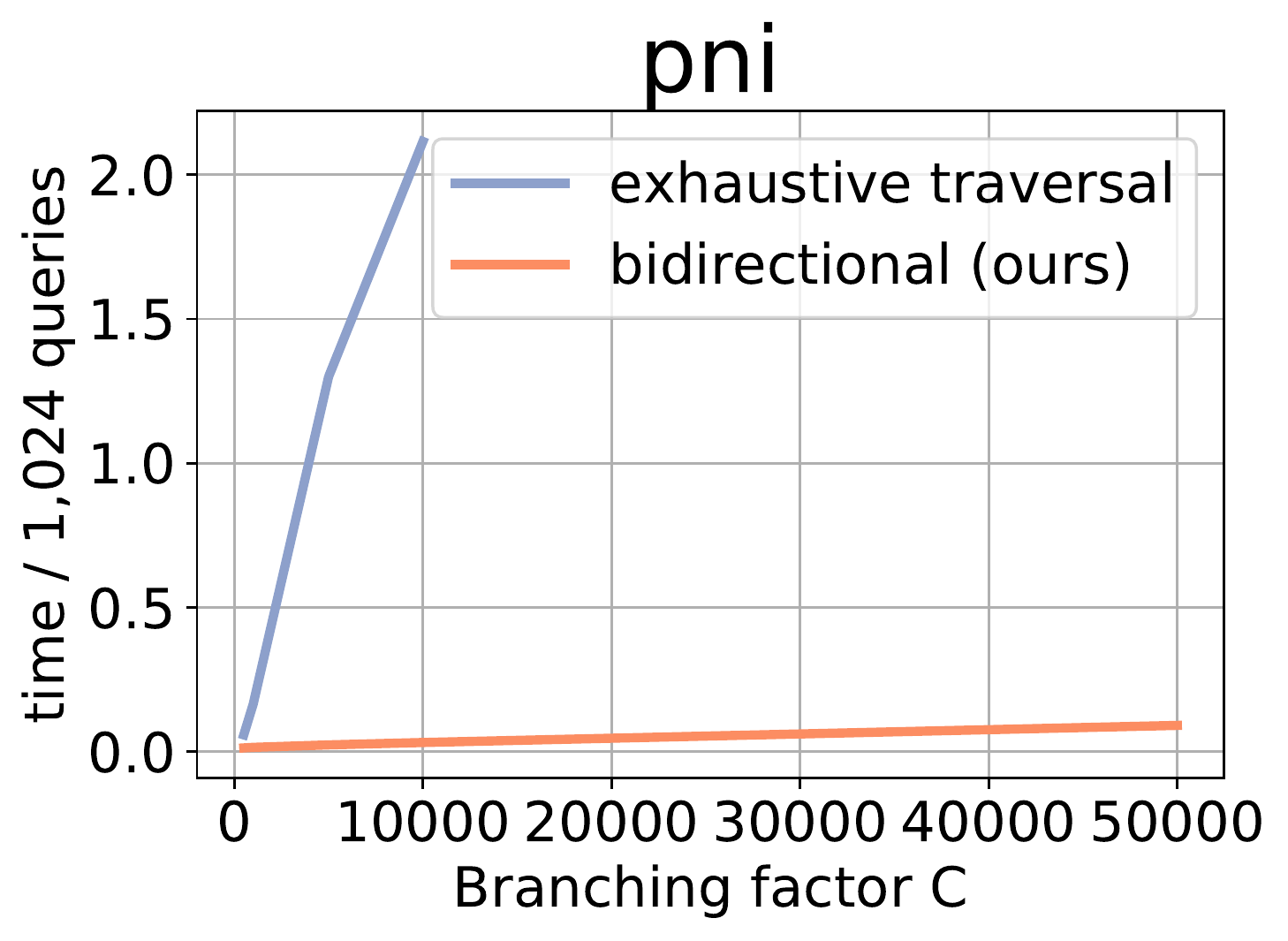}
\end{tabular}
\vspace{-1mm}
\caption{Speed-up of our bidirectional sampler over a na\"ive sampler that performs KG traversal, on 2p, ip and pni query structures (\Figref{fig:query_tree}). Our bidirectional sampling is significantly more efficient than the baseline.
\label{fig:sampler_speed}}
\end{figure*}

\begin{table}[t]
\centering
\caption{Runtime performance on Freebase KG. *Results taken from ~\citet{mohoney2021marius} with the same GPUs. \label{tab:link_prediction}
}
\vspace{-3mm}
\resizebox{0.48\textwidth}{!}{%
    \begin{tabular}{c|c|ccc}
    \toprule
    \multirow{2}{*}{System} &  \multicolumn{4}{c}{Epoch Time (s)} \\
    & 1-GPU & 2-GPU & 4-GPU & 8-GPU  \\
    \hline
    Marius~\citep{mohoney2021marius}* & 727 &  - & - & - \\
    DGL-KE~\citep{zheng2020dgl}* & - & 1068 & 542 & 277 \\
    PBG~\citep{lerer2019pytorch}* & 3060 & 1400 & 515 & 419 \\
    \modelshort{} & 760 & 411 & 224 & 121 \\
    \bottomrule
    \end{tabular}
}
\end{table}

\begin{figure*}[t]
\centering
\begin{minipage}{0.615\textwidth}
\includegraphics[width=0.495\textwidth]{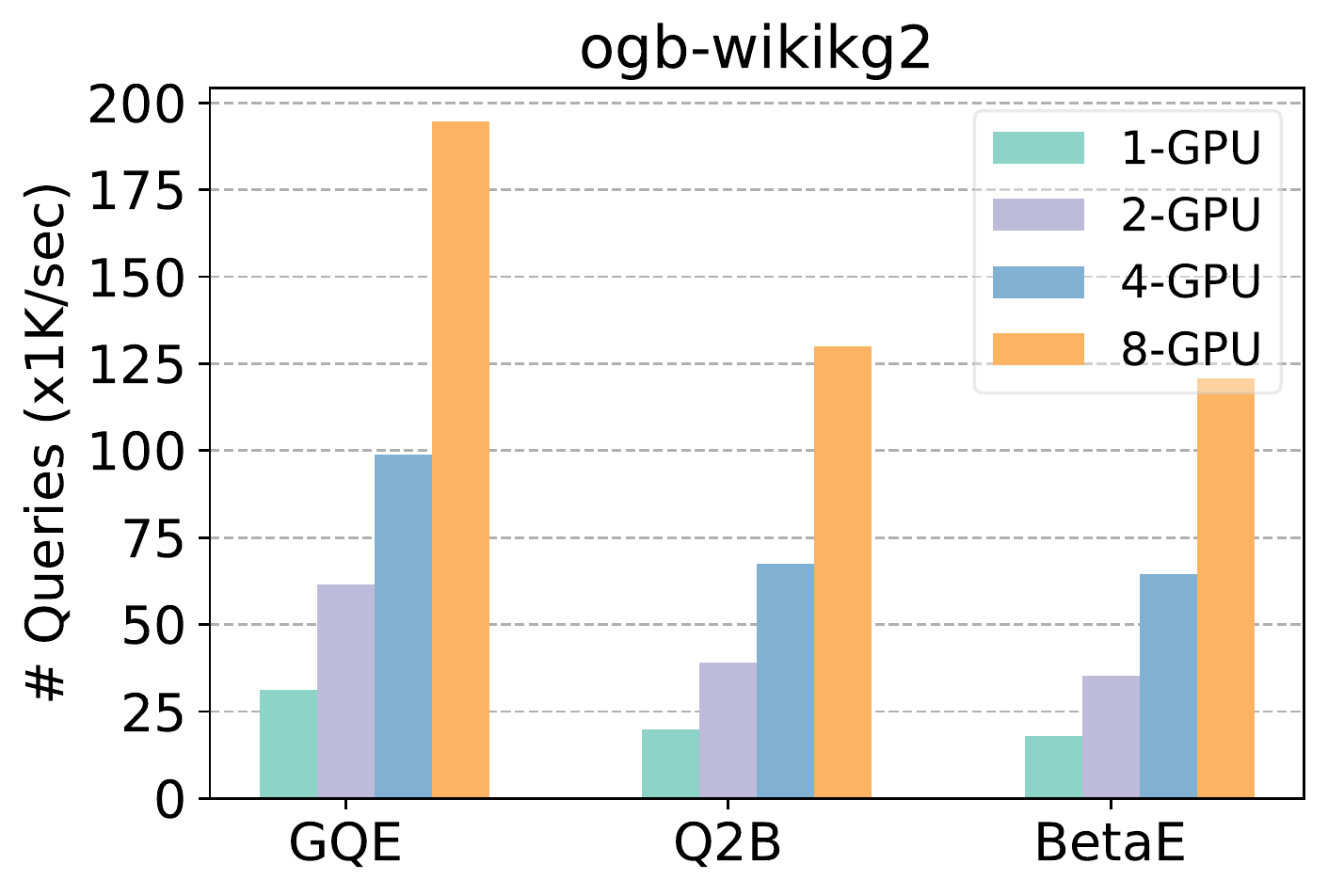}
\includegraphics[width=0.495\textwidth]{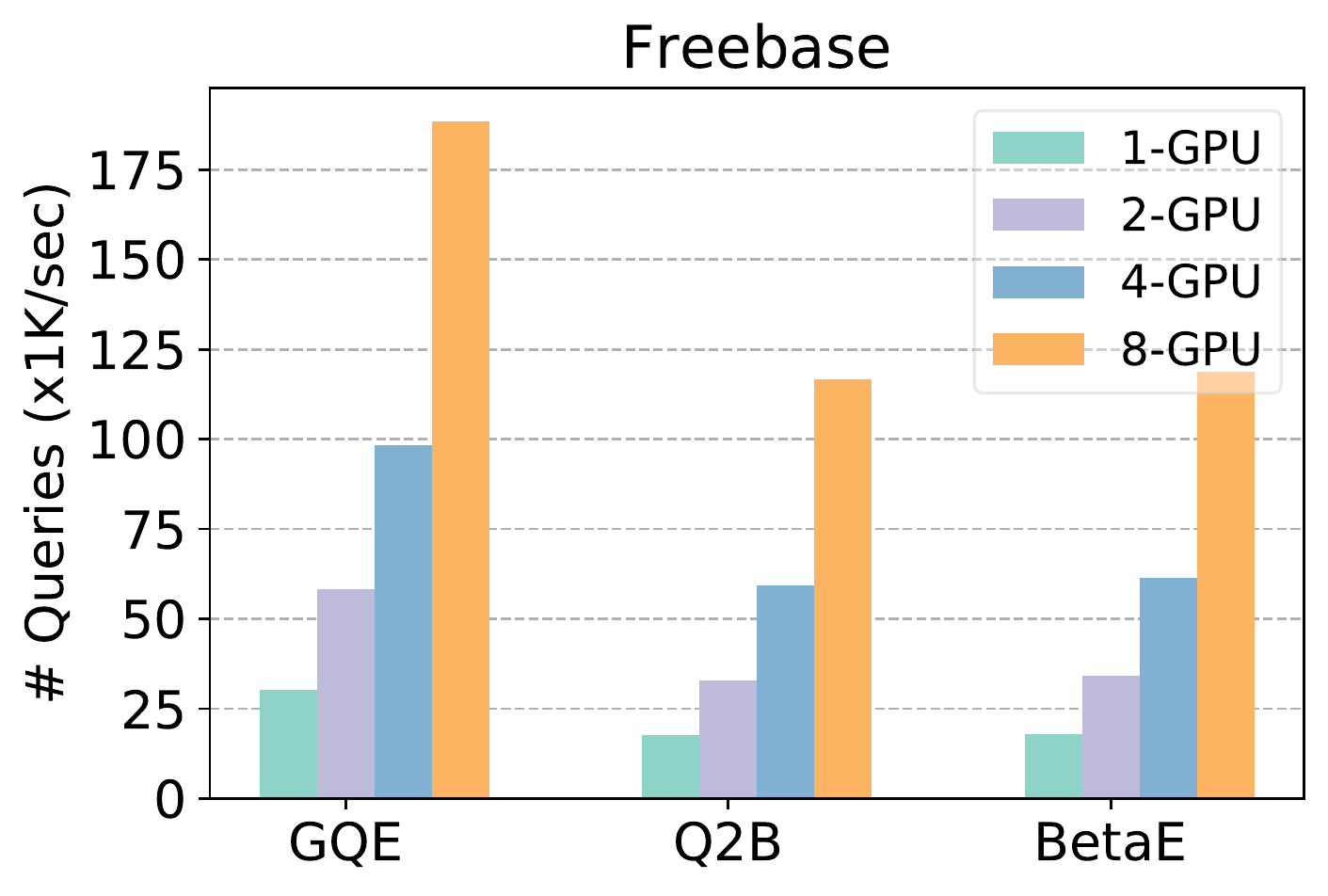}
\caption{\modelshort enjoys almost linear speed-up with respect to the number of GPUs, on both ogbl-wikikg2 and Freebase KGs, and across multiple embedding methods GQE, Q2B and BetaE. \label{fig:multigpu}}
\end{minipage}
\hfill
\begin{minipage}{0.38\textwidth}
\centering
\captionof{table}{Speed and GPU memory of \modelshort on the three large KGs with embedding dimension $=400$ and per device batch size $=512$. Our system design enables the (almost) graph-size agnostic speed and GPU memory usage.
\label{tab:largekgspeed} }
\vspace{-7mm}
\resizebox{0.98\textwidth}{!}{%
\begin{tabular}{c|c|c|cc|cc|cc}
\toprule 
Model & Dataset & Queries (x1K/Sec) & GPU Mem (MB) \\
\hline
\multirow{3}{*}{BetaE} &  FB400k & 123  & 1,872 \\
 & ogbl-wikikg2 & 121 & 1,860 \\
 & Freebase &  118 & 2,014 \\
\hline
\multirow{3}{*}{Q2B} &  FB400k & 135 & 1,796 \\
 & ogbl-wikikg2 & 130 & 1,776 \\
 & Freebase & 116 & 2,382 \\
 \hline
\multirow{3}{*}{GQE} &  FB400k & 189 & 1,726 \\
 & ogbl-wikikg2 & 195 & 1,750\\
 & Freebase & 188 & 2,056 \\
\bottomrule
\end{tabular}
}%
\end{minipage}
\end{figure*}

\section{Experimental results}\label{sec:exp}

Here we evaluate \modelshort on KG completion and multi-hop reasoning tasks on KG. The task is to answer multi-hop complex logical queries on KGs using various query embeddings. Besides the three small KGs used in prior works~\cite{ren2020query2box,ren2020beta}, we propose a novel set of multi-hop reasoning benchmarks on three extremely large KGs with more than 86 million nodes and 338 million edges. We first demonstrate the scalability of \modelshort on query sampling, multi-gpu speedup, GPU utilization over the three large KGs where all existing implementations fail. Additionally, we compare our single-hop KG completion runtime with state-of-the-art frameworks DGL-KE~\cite{zheng2020dgl}, Pytorch-Biggraph (PBG)~\cite{lerer2019pytorch} and Marius~\cite{mohoney2021marius}. Then we evaluate the end-to-end training performance of \modelshort on multi-hop reasoning over small as well as large KGs. This includes (1) calibration of the performance of three prior works GQE~\cite{hamilton2018embedding}, Q2B~\cite{ren2020query2box} and BetaE~\cite{ren2020beta} using \modelshort framework on the three small KGs; (2) a thorough evaluation of the query embedding models on the three large KGs where prior implementation~\cite{kgreasoning} is not applicable.

\subsection{Experimental setup}

\textbf{Task setup.} Following the standard experimental setup~\citep{ren2020query2box}, given an incomplete KG, the goal is to train query embedding methods to discover missing answers of complex logical queries. The detailed task setup and training/validation/test split can be found in \appref{app:task-setup}.
Following the standard evaluation metrics, we adopt mean reciprocal rank (MRR) with the filtered setting as our metric. 
The detailed procedure of calculating the metrics can be found in~\appref{app:metrics}.

\textbf{Datasets.}
We used the three datasets FB15k, FB15k-237, NELL from \citet{ren2020beta}. These three KGs are small-scale with at most 60k entities.
To create a set of large-scale multi-hop KG reasoning benchmarks, we further sample queries on three large KGs: FB400k, ogbl-wikikg2 and Freebase. 
Ogbl-wikikg2 is a KG from the Open Graph Benchmark~\cite{hu2020open}. 
FB400k is a subset of Freebase~\cite{bollacker2008freebase} which is derived based on a knowledge graph question answering dataset ComplexWebQuestion (CWQ)~\cite{talmor2018web}. 
We further look at the complete Freebase KG used in DGL-KE. 
For FB15k, FB15k-237 and NELL, we directly take the validation and test queries from \citet{ren2020beta}.
For ogbl-wikikg2 and Freebase, we randomly sample validation and test queries using the official edge splits. We consider the same 14 query structures proposed in BetaE~\cite{ren2020beta}. For FB400k, we directly take the SPARQL annotations of the validation and test questions in the CWQ as our validation and test queries.
Statistics of all the datasets can be found in Table \ref{tab:meta_kg}. The KGs we use here are up to 1500$\times$ larger than those considered by prior work.

\textbf{Software and training details.}
We implement all models in Python 3.8 using Pytorch 1.9~\citep{paszke2019pytorch} with customized CUDA ops, and the samplers in C++ with multithreading. The machine has 8 NVIDIA V100 GPUs, 96x 2.00 GHz CPUs and 600GB RAM. All models have same embedding dimension for fair comparison.

\begin{figure}[t]
\centering
\includegraphics[width=0.48\textwidth]{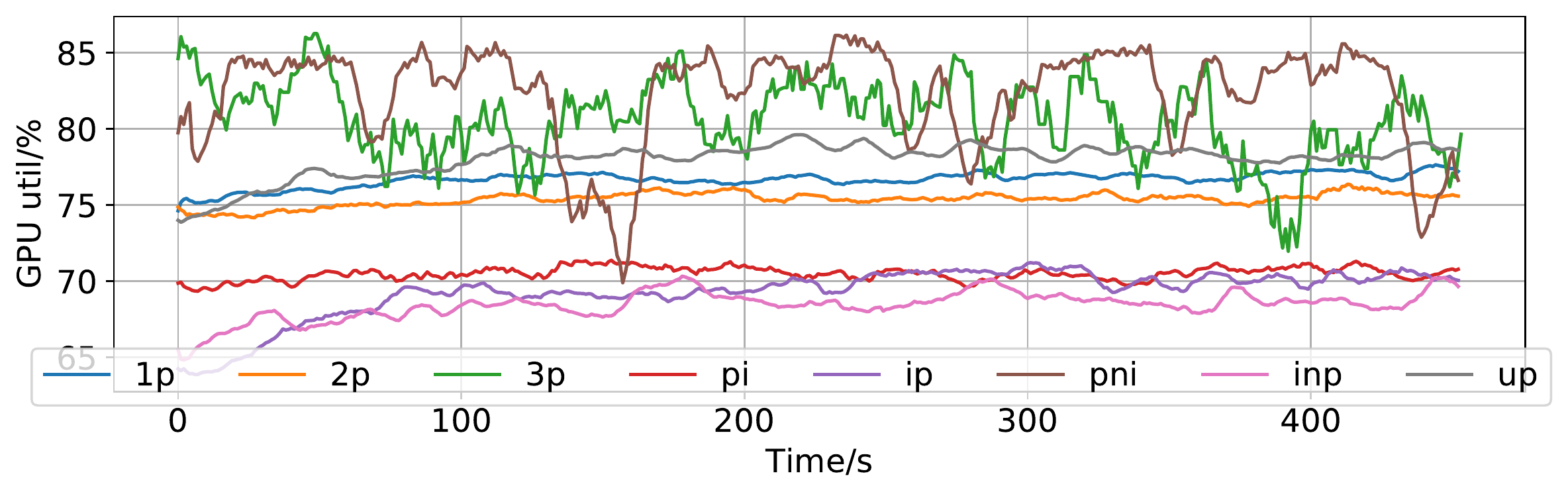}
\vspace{-4mm}
\caption{GPU utilization with different query structures on Freebase with BetaE~\citep{ren2020beta}. \label{fig:gpu_util} }
\end{figure}

\begin{figure*}[h!]
\centering
\begin{minipage}{0.31\textwidth}
\centering
\captionof{table}{MRR results (\%) on large KGs with different methods all trained using \modelshort. 
Results for each query structure can be found in~\appref{app:detailedmrr}.
\label{tab:mergedmrr} 
}
\resizebox{1.0\textwidth}{!}{%
\begin{tabular}{c|ccc}
\toprule
{\small Dataset \textbackslash\: Model} & GQE & Q2B & BetaE\\
\hline
FB400k & 36.02 & {\bf 51.74} & 50.50 \\
ogbl-wikikg2 & 32.91 & 41.88 & \textbf{44.42} \\
Freebase & 80.71 & \textbf{85.67} & 84.33 \\
\bottomrule
\end{tabular}
}%
\end{minipage}
\hfill
\begin{minipage}{0.67\textwidth}
    \centering
    \includegraphics[width=1.0\textwidth]{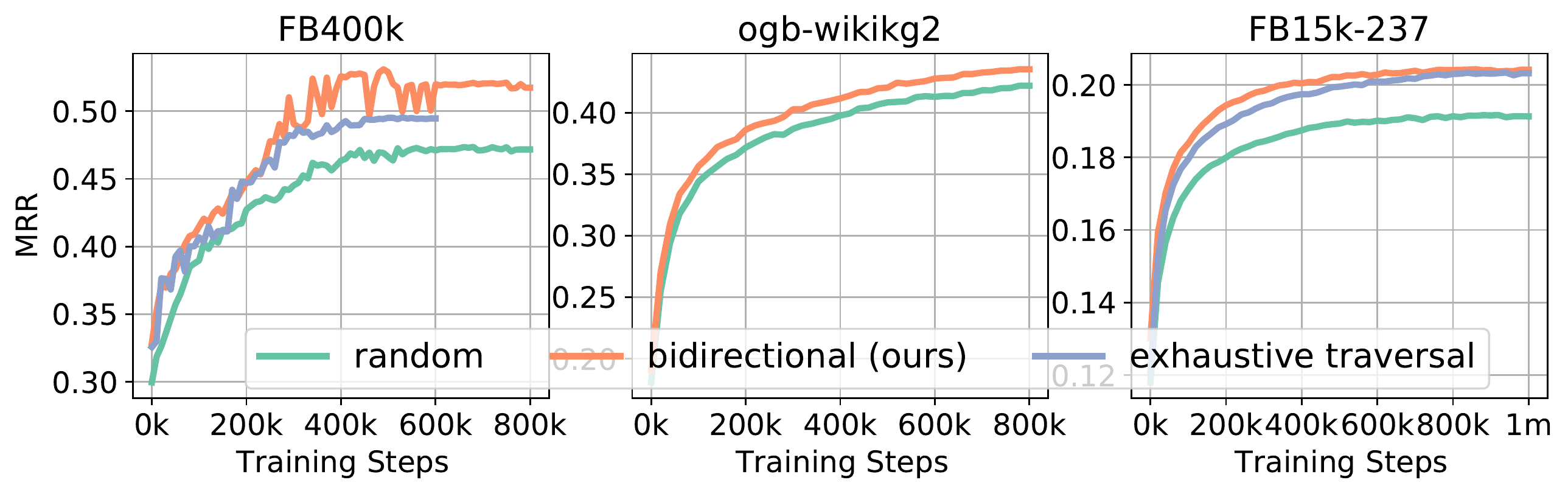}
    \vspace{-7mm}
    \caption{Accuracy (MRR) of different Q2B models trained using different samplers. We observe that our bidirectional sampler leads to more accurate models.
\label{fig:sampling_result} }
\end{minipage}
\end{figure*}

\subsection{Scalability}
\textbf{Speedup of bidirectional sampler.}
We verify the speed-up of the proposed bidirectional sampler over na\"ive exhaustive traversal one in \Figref{fig:sampler_speed} (more results on other query structures can be found in \appref{app:more_exp_results}). We test with different query structures where the bidirectional sampler is expected to achieve a square root reduction of the computation cost compared with traversal. We vary the constant $C$, \ie, the maximum expansion of each relation, and plot the time for sampling a minibatch of 1024 queries as the function of $C$ using the largest Freebase KG. We can see the exhaustive traversal approach runs out of the time limit quickly when $C$ grows, while our proposed sampler stays significantly more efficient across all query structures (\Figref{fig:query_tree}).

\textbf{GPU memory, utilization and end-to-end training speed.}
Here we show the scalability of \modelshort on all six KGs. We first compare \modelshort with prior implementation~\cite{kgreasoning} on the three small benchmark datasets created in \citet{ren2020beta}. As shown in the last two columns in Table~\ref{tab:calibrate}, \modelshort significantly improves the efficiency of end-to-end training of various query embeddings since \modelshort adopts a query sampling scheme that shares the negative samples for a sampled batch of queries. Specifically, \modelshort increases the speed by 119.4\% and reduces GPU memory usage by 30.6\% on average. Then we scale query embeddings up to the three large KGs. As shown in Table~\ref{tab:largekgspeed},  given the same embedding dimension and batch size, our system design allows for (almost) graph-size agnostic speed and GPU-memory usage across all methods.

In \Figref{fig:gpu_util} we further show the GPU utilization of BetaE model with different multi-hop query structures on Freebase. We plot the average utilization of 8 GPUs with smoothing of 10 seconds. More complex queries like \texttt{3p} and \texttt{pni} would have higher fluctuation due to the variance of sampling time. However as these queries also require more neural ops which in turn brings up the GPU utilization. Generally our system is able to keep a high GPU utilization for all these query structures.

\textbf{Multi-GPU speed up.}
We illustrate the speed-up of training on $\{$1, 2, 4, 8$\}$ GPUs with different methods on ogbl-wikikg2 and Freebase datasets in~\Figref{fig:multigpu}. Overall the speed grows almost linearly w.r.t. the number of GPUs, which shows the effectiveness of our asynchronous training and the communication overhead is negligible. Also the computationally heavier approaches like BetaE, which requires calculating KL-divergence between high-dimensional Beta distributions as well as its derivatives, benefit more from multi-GPU on \modelshort. 

\textbf{KG completion runtime.}
Here we compare the single-hop link prediction (KG completion) runtime performance with state-of-the-art large-scale KG frameworks including Marius, DGL-KE and PBG. We report the results for ComplEx model with 100 embedding dimension on Freebase. We use the same multi-GPU V100 configuration as in Marius, and the current official release versions are also the same. So for baseline results we reuse the table 7 from~\citet{mohoney2021marius}. As shown in Table~\ref{tab:link_prediction}, \modelshort achieves significantly faster runtime in 1-GPU setting than PBG, while being slightly slower than Marius. Also it scales better than the other systems, while Marius does not officially support multi-GPU parallel training functionality at the current stage. Given that \modelshort adopts the design choice of synchronized gradient update for dense parameters, and we do not partition the graph for the sake of multi-hop reasoning, it is nontrivial to be still comparable in the single-hop case.

\subsection{Predictive performance}
\textbf{Calibration on small KGs.}
We first calibrate our \modelshort system for GQE, Q2B, and BetaE on the three small benchmark datasets created in BetaE. As shown in Table~\ref{tab:calibrate}, \modelshort achieves overall comparable performance as the models trained on a fixed set of training queries using a single GPU. Specifically, \modelshort achieves comparable results for GQE and BetaE respectively but is able to further improve the performance of Q2B on FB15k with 3.54\% increase in MRR.

\textbf{Query answering on large KGs.}
We also benchmark the embedding models on the three large KGs FB400k, ogbl-wikikg2 and Freebase. As shown in \tabref{tab:mergedmrr}, on ogbl-wikikg2, BetaE performs the best among all the other baselines. On both FB400k and Freebase, box embedding model Q2B achieves the best results. For all these methods, the baseline implementation~\cite{kgreasoning} cannot scale to such massive KGs due to the short of GPU memory and computationally expensive exhaustive query sampling. 
Our system \modelshort easily scales query embeddings to these large KGs using an asynchronous design with sparse embedding and optimizer. 
The three KGs serve as an important benchmark for future multi-hop KG reasoning models.

\textbf{Performance with different samplers.}
We compare the performance of Q2B trained with different samplers, \ie, the na\"ive sampler (\texttt{exhaustive traversal}), bidirectional sampler (\texttt{bidirectional}) and randomly sampling KG entities as negative answers (\texttt{random}) on FB15k-237, FB400k and ogbl-wikikg2. As shown in \Figref{fig:sampling_result}, random sampling performs the worst since random negative sampling does not guarantee that the sampled entities are truly the negative (non-answer) entities. \texttt{bidirectional} performs comparable or even better than \texttt{exhaustive traversal}, as it can choose much larger $C$ during query sampling. Note that \texttt{exhaustive traversal} is very slow on large graphs, and it requires months for the baseline implementation with exhaustive traversal sampler to train a model with the same number of queries on ogbl-wikikg2.
\section{Conclusion}
We present \modelshort, the first general framework for both single- and multi-hop reasoning 
that scales up a plenty of different embedding methods with multi-GPU support to KGs with 86M nodes and 338M edges. It performs the algorithm-system co-optimization for scalability. Our work can also serve as the benchmark for future research on large-scale multi-hop KG reasoning. 

\section*{Acknowledgements}
We thank Theo Rekatsinas, Rok Sosic, Xikun Zhang, Serena Chang, Michael Xie and Sharmila Nangi for providing feedback on our manuscript, Jason Mahoney and Roger Waleffe for discussions on Marius, Matthias Fey for discussions on sparse embeddings.
We also gratefully acknowledge the support of 
DARPA under Nos. HR00112190039 (TAMI), N660011924033 (MCS);
ARO under Nos. W911NF-16-1-0342 (MURI), W911NF-16-1-0171 (DURIP);
NSF under Nos. OAC-1835598 (CINES), OAC-1934578 (HDR), CCF-1918940 (Expeditions), IIS-2030477 (RAPID),
NIH under No. R56LM013365;
Stanford Data Science Initiative, 
Wu Tsai Neurosciences Institute,
Chan Zuckerberg Biohub,
Amazon, JPMorgan Chase, Docomo, Hitachi, Intel, JD.com, KDDI, NVIDIA, Dell, Toshiba, Visa, and UnitedHealth Group. 
Hongyu Ren is supported by the Masason Foundation Fellowship and the Apple PhD Fellowship.
Jure Leskovec is a Chan Zuckerberg Biohub investigator.

\bibliography{bibfile}
\bibliographystyle{mlsys2022}

\newpage
\onecolumn
\appendix
\begin{center}
\begin{huge}
\textbf{Appendix}
\end{huge}
\end{center}

\section{Query computation plan}\label{app:computegraph}

As shown in~\Figref{fig:query}, the computation plan of a query consists of nodes $\Vcal_q \cup \{V_1, \dots, V_k, V_?\}$. Note each node of the computation plan corresponds to \emph{a set of entities} on the KG. The edges on the computation plan represent a logical/relational transformation of this set:
\vspace{-1mm}
\begin{enumerate}[noitemsep,topsep=0pt,parsep=1pt,partopsep=0pt, leftmargin=*]
    \item \textbf{Relation Projection:} Given a set of entities $S \subseteq \Vcal$ and relation type $r \in \Rcal$, compute adjacent entities $\cup_{v \in S} A_r(v)$ related to $S$ via $r$: $A_r(v) \equiv \{v^{\prime} \in \Vcal: \ (v, r, v^{\prime})\in\Ecal\}$.
    \item \textbf{Intersection:} Given sets of entities $\{ S_1, S_2, \ldots, S_n\}$, compute their intersection $\cap_{i = 1}^n S_i$.
    \item \textbf{Complement/Negation:} Given a set of entities $S \subseteq \Vcal$, compute its complement $\overline{S} \equiv \Vcal \:\backslash\: S$.
\end{enumerate}
\vspace{-1mm}
According to the De Morgan's laws,  
$\cup_{i=1}^n S_i$ is equivalent to $\overline{\cap_{i=1}^n \overline{S}}$, union operation can be replaced with three negation and one intersection operations. 

\section{Multi-hop reasoning models}\label{appendix:models}

Our \modelshort covers six models with three published works, \ie,  GQE~\cite{hamilton2018embedding}, Q2B~\cite{ren2020query2box} and BetaE~\cite{ren2020beta}, as well as three new methods based on single-hop link prediction models. 
In order to perform logical reasoning in the embedding space, all methods design a projection operator $\Pcal$ and intersection operator $\Ical$. As introduced in ~\Secref{sec:background}, $\Pcal$ represents a mapping from a set of entities (represented by an embedding) to another set of entities (also represented by an embedding) with one relation, \ie, $\Pcal: \mathbb{R}^d\times\Rcal\to\mathbb{R}^d$, assuming the embedding dimension is $d$. The $\Ical$ takes as input multiple embeddings and outputs the embedding that represents the intersected set of entities: $\Ical: \mathbb{R}^d\times\dots\times\mathbb{R}^d\to\mathbb{R}^d$. Different models may have different instantiations of these two operators.

GQE \citep{hamilton2018embedding} is the first method that handles conjunctive queries. GQE embeds a query $q$ to a point in the vector space by iteratively following the computation plan of the query. 
The distance function between the query embedding and the node embedding is the $L_2$ distance: $\texttt{Dist}(\Em{q},\Em{v})=\|\Em{q}-\Em{v}\|_2$. 
Query2box (Q2B)~\cite{ren2020query2box} proposes to embed queries as hyperrectangles with a center embedding and an offset embedding. Q2B simulates set intersection using box intersection. The distance function in Q2B is a modified L1 distance: a weighted sum of the in-box distance and out-box distance.
BetaE~\cite{ren2020beta} further proposes to embed queries as multiple independent Beta distributions so that it can faithfully handle conjunction and negation operation, and thus disjunction by the De Morgan's laws in distribution space with $KL$-divergence as distance.

Based on the existing methods, we also extend several link prediction models to perform multi-hop reasoning, including RotatE~\cite{sun2019rotate}, DistMult~\cite{yang2014embedding} and ComplEx~\cite{trouillon2016complex}. 

RotatE models a relation by rotation in the complex space, which is naturally compositional. We extend RotatE to RotatE-m for multi-hop reasoning. The relation projection $\Pcal$ is defined as
$\Pcal(\Em{q},\Em{r})=\Em{q}\circ\Em{r}, |\Em{r}[i]|=1$. For the intersection operator, we follow GQE that uses a DeepSets architecture taking multiple complex vectors as input and outputing the intersected result. 
The same distance function in \citet{sun2019rotate} are used, $\texttt{Dist}(\Em{q},\Em{v})=\|\Em{q}-\Em{v}\|$.

Following the design of multihop DistMult proposed in the GQE~\cite{hamilton2018embedding}, we also extend ComplEx to perform multi-hop reasoning. The relation projection is defined as
$\Pcal(\Em{q},\Em{r})=\Em{q}\circ\Em{r}$. And we follow GQE and RotatE-m using the same DeepSets architecture for the intersection operator. The distance function (negative of similarity score) is defined as: $\texttt{Dist}(\Em{q},\Em{v})=-\text{Re}(\langle\Em{q},\overline{\Em{v}}\rangle)$. One key problem of this extension to these semantic matching models is that since the embeddings are not normalized, the similarity score tends to explode especially in the presence of multi-hop relation projection. Thus, we normalize the query embeddings after each step of relation projection or intersection and propose DistMult-m and ComplEx-m. For DistMult-m, we directly perform $L_2$ normalization: $\Em{q}=\frac{\Em{q}}{\|\Em{q}\|_2}$; for ComplEx-m, we normalize the real and imaginary parts separately to the unit hypersphere: 
$\text{Re}(\Em{q})=\frac{\text{Re}(\Em{q})}{\|\text{Re}(\Em{q})\|_2}, \text{Im}(\Em{q})=\frac{\text{Im}(\Em{q})}{\|\text{Im}(\Em{q})\|_2}$.

\section{Reverse directional sampling}
\label{app:reverse_sample}
Following prior work \citep{ren2020query2box,ren2020beta}, we use reverse directional sampling to construct queries from a KG. The overall instantiation process corresponds to a depth-first search where at each step we aim to ground a node/edge on the query structure with an entity/relation from the KG. Here we use one example to illustrate the whole idea. As shown in \Figref{fig:query}, if we aim to instantiate a \texttt{ip} query from the query structure, then we start the instantiation process from the root node, where we randomly sample an entity from the KG, here being \texttt{Neal}. Then following the query structure, we aim to ground the edge that points to the root in the query structure, and we sample a relation type from the KG that points to the entity \texttt{Neal}, \eg, \texttt{Co-author}. Then we ground the next node, which has the relation \texttt{Co-author} with \texttt{Neal}, \eg, \texttt{Bengio}. In the next step, since the edge is a logical operation \texttt{Intersection}, we may directly ground the next node with the same entity \texttt{Bengio}, and we sample another relation on the KG the relates to \texttt{Bengio}, \eg, \texttt{Win}, and finally reaches the anchor entity (leaf) by sampling an entity on KG that has the relation \texttt{Win} with \texttt{Bengio}, \eg, \texttt{Turing Award}. The overall complexity of this process is linear with respect to the number of hops of a query (structure).

\textbf{Dynamic programming for optimal node cut.}
\label{app:cut_dp}
As the computation plan of $q$ is a tree, we propose to solve the above optimization problem in \eqnref{eq:optimal_sheduling} with dynamic programming (DP). Before presenting the algorithm, we first need to understand the cost of each operation, so as to only model the dominating cost in the dynamic programming. 

We consider the following operations: 
\begin{itemize}[noitemsep,topsep=0pt,parsep=0pt,partopsep=0pt, leftmargin=*]
    \item \textit{Relation projection}: this operation enlarges the current set of entities by a factor of $C$ (the maximum node degree) in the worst case. Thus the total cost would grows exponentially with the number of relation projection operations in a reasoning path.
    \item \textit{Intersection / Union}: if we maintain the set of entities as a sorted list, intersection / union of the two sets only takes linear time w.r.t the number of entities in both sets. Thus it will not be the limiting factor in the overall computation cost if we only merge a constant number of sets together. 
    \item \textit{Negation / Complement}: The computation cost of a single set complement operation $\Ocal(|\Vcal|)$, \ie, the total number of entities in the KG. However, we can delay the complement operation to the next step on the computation plan, \ie, perform complement+union/intersection simultaneously. This reduces the complexity from $\Ocal(\Vcal)$ to that of an intersection operation. For example in $(\neg a) \wedge b$, instead of first finding the complement of $a$ (of complexity $|\Ocal(\Vcal)|$) and then do $\wedge$ (of complexity $\Ocal(|\Vcal-a|+|b|)$), we can directory do set difference $b-a$ (of complexity $\Ocal(|a|+|b|)$.
    The only exception is the relation projection after the negation immediately. However there are other equivalent forms for this expression and thus we explicitly exclude such possibility in constructing the query structure.
\end{itemize}
With the analysis above, we can see the only thing we need to focus is the maximum number of relation projections in any reasoning path (\ie, a path that connects a leaf/anchor entity $v\in \Vcal_q$ and the root/answer entity $V_?$).

We first define a set of functions:
\begin{itemize}[noitemsep,topsep=0pt,parsep=0pt,partopsep=0pt, leftmargin=*]
	\item $u(v)$: number of relation projections in the path from node $v$ to root $V_?$;
	\item $s(v)$: the maximum length of path from $v$ to any anchors. The ``length'' is measured by the number of relation projections on that path.
	\item $o(v)$: the optimal cost of resolving all the reasoning paths that includes $v$ in the best plan. Note this cost only cares the dominating one under the big-O notation, not the cost of entire search/reasoning.
\end{itemize}
Note that the ``cost'' are measured in the logarithmic scale, as we only care the dominating order of the polynomials in the complexity calculation. 
We use $p(v)$ to denote the parent of node $v$, and $ch(v)$ to denote the set of children. When $v$ only has one child node, then we overload $ch(v)$ to denote that specific child. Then we can work on the recursion as below:
\begin{eqnarray}
	u(v) &=& u(p(v)) + \text{IsRel}(v \rightarrow p(v)) \nonumber \\
	s(v) &=& \begin{cases}
     0, \quad \text{if }v \in \Vcal_q \\
     \max_{z \in ch(v)} s(z), \quad \\
     \quad \quad \text{if edges between $v$ and $ch(v)$ are $\wedge$ or $\vee$} \nonumber \\
    s(ch(v)) + \text{NotNeg}(ch(v) \rightarrow v), \quad \text{else} \\
            \end{cases} \\
     o(v) &=& \begin{cases}
    u(v), \quad \text{if }v \in \Vcal_q \nonumber \\
     {\displaystyle \min\cbr{\max_{z \in ch(v)} o(z), \max\cbr{u(v), s(v)}}}, \quad \text{else} \\
            \end{cases}
\end{eqnarray}

$\text{IsRel}(v\to p(v))$ returns 1 if the edge between $v$ and $p(v)$ represents a relation projection and 0 otherwise; $\text{NotNeg}(ch(v)\to v)$ returns 1 if the edge between $ch(v)$ and $v$ is not negation and 0 otherwise.

After solving the above DP, we can construct the node cut from solution $o(\cdot)$ in a top-down direction:
\begin{itemize}[noitemsep,topsep=0pt,parsep=0pt,partopsep=0pt, leftmargin=*]
	\item If for any node $v$ we have $\max_{z \in ch(v)} o(z)$ larger than\\ $\max\cbr{u(v), s(v)}$, then we add $v$ to node cut;
	\item Otherwise, we do the check recursively for $z \in ch(v)$. 
\end{itemize}

The above procedure works in a linear runtime w.r.t. the size of $q$, which is good enough for some large queries containing hundreds of operations. Example query structures and their corresponding optimal node cuts are shown in \Figref{fig:query_tree}. Note the structures considered in the current literature are small enough to find the optimal cut with the brute force algorithms. But our DP can greatly improve the efficiency when the query structures are large enough in certain applications.

\section{Comparison with neural link predictor \cite{arakelyan2020complex}}\label{app:link-predictor}
Here we discuss the neural link predictor \cite{arakelyan2020complex}, which uses fuzzy logic to answer complex queries. It either requires online optimization for test query answering or traverses the KG with beam search, which is hard to scale to large KGs with their own unique challenges. In this paper, we focus on scaling up query embedding methods and leave scaling fuzzy-logic-based methods as future work.

\section{Detailed task setup}\label{app:task-setup}
Following the standard experimental setup~\citep{ren2020query2box,ren2020beta}, the goal is to evaluate whether the trained model is able to discover the missing answers of a query on an incomplete KG.
Following the standard evaluation metrics, we adopt mean reciprocal rank (MRR) with the filtered setting as our metric. 
Given the training, validation, test splits of the edges $\Ecal_\text{train}$, $\Ecal_\text{valid}$, $\Ecal_\text{test}$ in a KG, we create three subgraphs $\Gcal_\text{train}=(\Vcal, \Ecal_\text{train}, \Rcal)$, $\Gcal_\text{valid}=(\Vcal, \Ecal_\text{train}\cup\Ecal_\text{valid}, \Rcal)$, $\Gcal_\text{test}=(\Vcal, \Ecal_\text{train}\cup\Ecal_\text{valid}\cup\Ecal_\text{test},\Rcal)$. During training, we sample queries from the $\Gcal_\text{train}$ using our sampler as introduced in~\Secref{sec:sampler} and train each multi-hop reasoning model. During validation, we are given a set of evaluation queries. For each query, we may first traverse the training KG, $\Gcal_\text{train}$, to obtain its training answers $\Acal_q^{\Gcal_\text{train}}$, and then we further traverse the validation KG ,$\Gcal_\text{valid}$, to obtain the answers $\Acal_q^{\Gcal_\text{valid}}$. The evaluation setup is to evaluate the performance of the model on predicting all missing answers $v\in\Acal_q^{\Gcal_\text{valid}}\backslash\Acal_q^{\Gcal_\text{train}}$. During the test phase, we have the same setup, the only difference is that we evaluate whether the model can discover all missing answers $\Acal_q^{\Gcal_\text{test}}\backslash \Acal_q^{\Gcal_\text{valid}}$.

\section{Calculation of evaluation metrics}\label{app:metrics}
Given a test query $q$, we rank each missing answer $v\in\Acal_q^{\Gcal_\text{test}}\backslash \Acal_q^{\Gcal_\text{valid}}$ against the $\Vcal \backslash\Acal_q^{\Gcal_\text{test}}$, filtering the other answers. For large-scale graphs, comparing $v$ with all $\Vcal \backslash\Acal_q^{\Gcal_\text{test}}$ is impractical, thus, following standard practice in link prediction tasks~\citep{zheng2020dgl}, we randomly sample 1000 negative answers from $\Vcal \backslash\Acal_q^{\Gcal_\text{test}}$ for each query. Denote the rank of answer $v$ as $\text{Rank}(v)$, the metrics of a query $q$ can be calculated by:
\begin{align}
    &\text{Metrics}(q) = \frac{1}{|\Acal_q^{\Gcal_\text{test}}\backslash \Acal_q^{\Gcal_\text{valid}}|} \sum_{v \in \Acal_q^{\Gcal_\text{test}}\backslash \Acal_q^{\Gcal_\text{valid}}} f_{\text{metrics}}(\text{Rank}(v)), \nonumber
\end{align}
where $f_\text{metrics}(x)=\frac{1}{x}$ for MRR and $f_\text{metrics}(x)=1[x\leq k]$ for Hit@$k$. For a test set of queries, we average the metrics for each query in the set as the final performance of the model.

\section{Additional results on bidirectional rejection sampler}
\label{app:more_exp_results}

\begin{figure*}[t]
\includegraphics[width=0.325\textwidth]{speed-2p}
\includegraphics[width=0.325\textwidth]{speed-ip}
\includegraphics[width=0.325\textwidth]{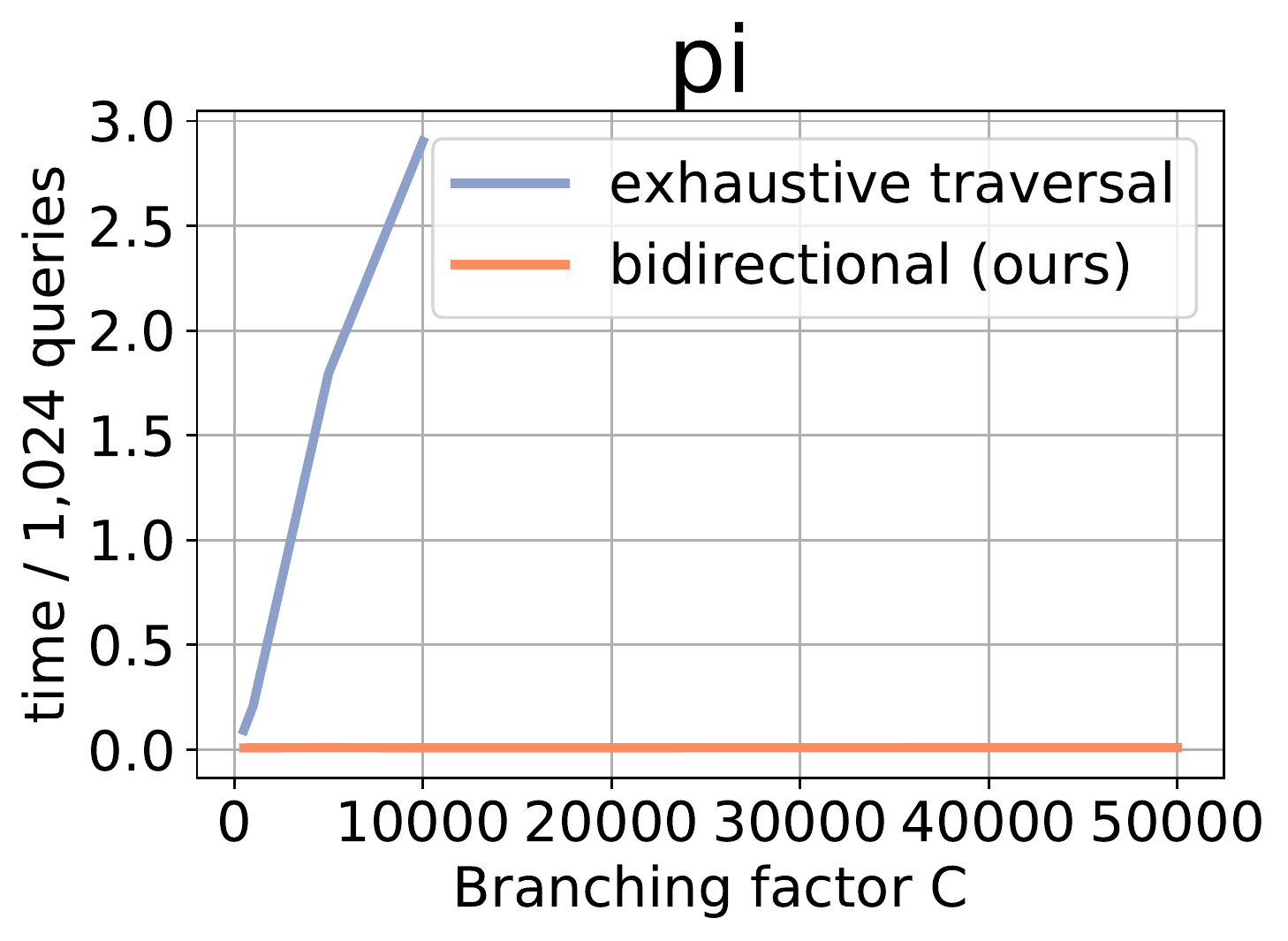}\\

\includegraphics[width=0.325\textwidth]{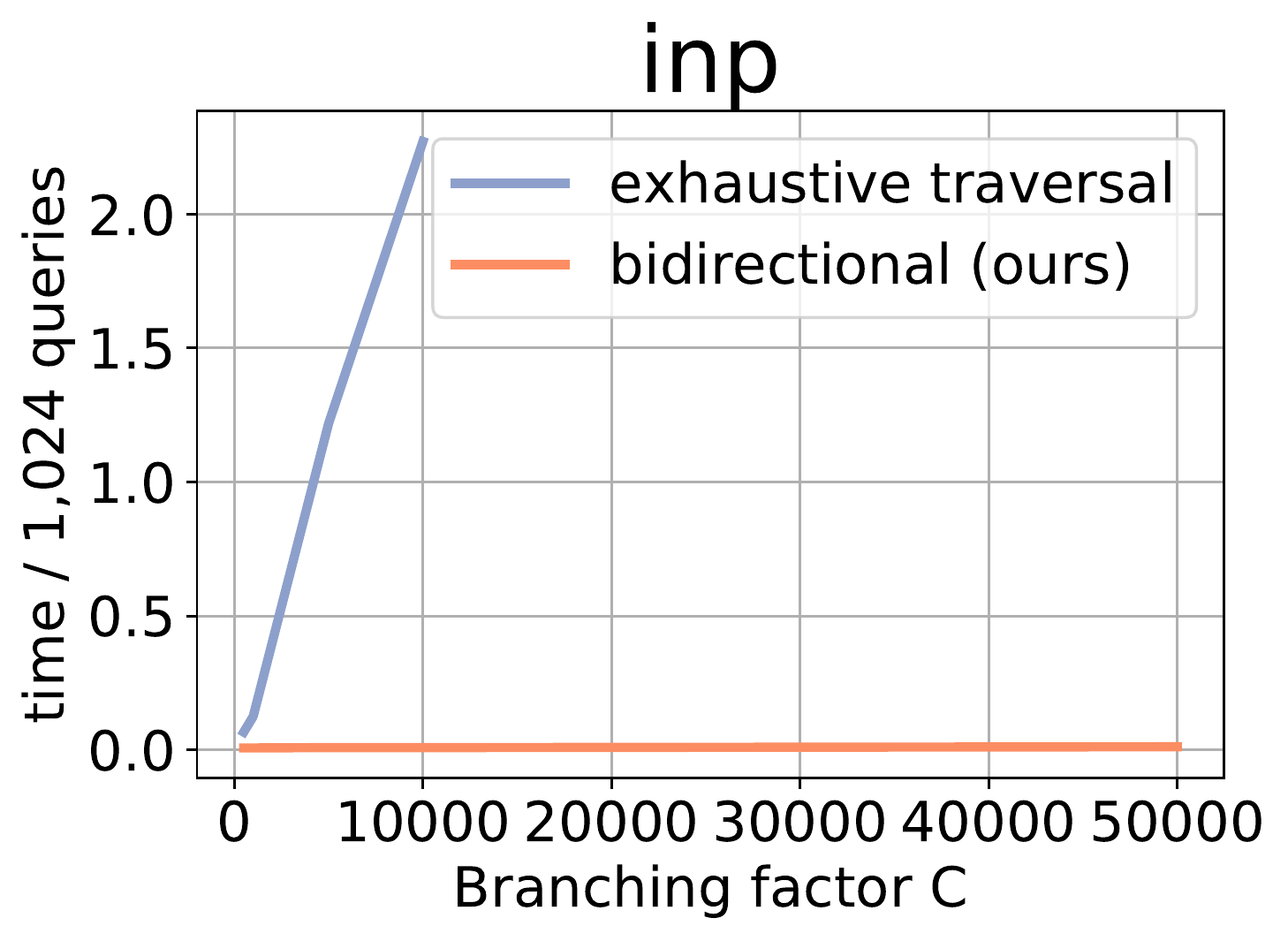}
\includegraphics[width=0.325\textwidth]{speed-pni}
\includegraphics[width=0.325\textwidth]{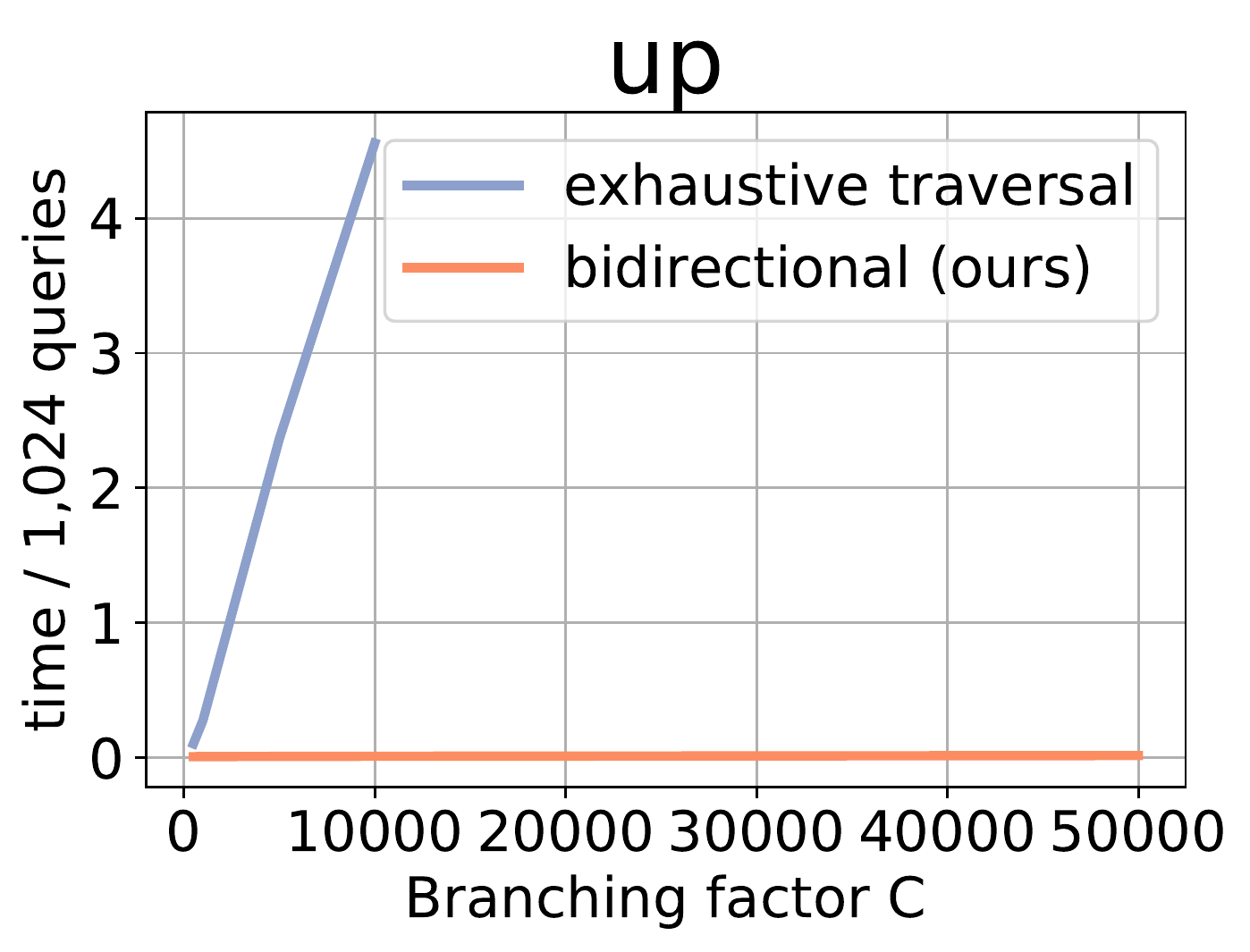}
\caption{Speedup of bidirectional rejection sampler over exhaustive search based sampler, on different query structures. \label{fig:more_sampler_speed}}
\end{figure*}

\Figref{fig:more_sampler_speed} shows the speed-up of bidirectional rejection sampler over exhaustive samplers on several more query structures. Ours is consistently significantly better than exhaustive traversal.

\section{Detailed MRR on large KGs}\label{app:detailedmrr}
Here we list the detailed MRR results (\%) of each reasoning model on large KGs in Table \ref{tab:large_kg_results}. For FB400k, we report the performance in Table \ref{tab:kgqa} on answering the SPARQL queries from the questions in Complex Webquestions dataset. The standard deviation of three runs on all query structures are less than 0.15.

\begin{table*}[t]
\centering
\caption{MRR results (\%) on large KGs with different methods. \label{tab:large_kg_results} }
\resizebox{\columnwidth}{!}{%
\begin{tabular}{c|c|ccc|cc|cc|c|c|c}
	\toprule
	Dataset & Model & 1p & 2p & 3p & 2i & 3i & pi & ip & 2u-DNF & up-DNF & avg \\
	\hline
\multirow{6}{*}{OGB} & Q2B & 39.32 & 33.95 & 33.71 & 48.69 & 61.21 & 43.76 & 36.39 & 43.23 & 36.63 & 41.88 \\
	& BetaE & 41.93 & 36.35 & 37.76 & 53.94 & 67.42 & 48.37 & 36.36 & 38.77 & 38.89 & 44.42 \\ 
	& GQE & 33.56 & 27.70 & 28.21 & 34.14 & 42.14 & 32.25 & 30.35 & 36.08 & 31.78 & 32.91 \\
	& RotatE & 28.28 & 24.37 & 25.10 & 44.54 & 56.95 & 37.94 & 23.80 & 34.50 & 31.14 & 34.07 \\
	& DistMult & 27.29 & 24.38 & 26.19 & 27.95 & 34.95 & 26.54 & 26.54 & 32.52 & 29.32 & 28.41 \\
	& ComplEx & 28.71 & 26.82 & 28.63 & 29.26 & 36.19 & 28.10 & 25.97 & 30.20 & 31.77 & 29.52 \\
	\hline
\multirow{6}{*}{Freebase} & Q2B & 84.58 & 80.36 & 78.68 & 92.41 & 93.96 & 86.80 & 84.84 & 92.06 & 77.33 & 85.67 \\
	& BetaE & 83.73 & 77.13 & 74.27 & 92.37 & 94.29 & 89.72 & 82.63 & 91.21 & 73.63 & 84.33 \\ 
	& GQE & 80.04 & 74.53 & 71.58 & 89.17 & 90.78 & 83.20 & 79.25 & 86.36 & 71.11 & 80.71 \\
	& RotatE & 81.20 & 75.49 & 73.37 & 85.95 & 89.14 & 82.97 & 76.08 & 88.00 & 72.70 & 80.55 \\
	& DistMult & 69.27 & 68.08 & 66.27 & 71.88 & 75.52 & 68.65 & 69.03 & 77.78 & 65.70 & 70.24\\
	& ComplEx & 72.47 & 64.65 & 59.47 & 78.51 & 84.50 & 65.68 & 72.08 & 77.26 & 60.66 & 70.59 \\	
	\bottomrule
\end{tabular}
}%
\end{table*}
\begin{table}[t]
\centering
\caption{Predictive performance on Complex WebQuestion test set. Hit@10 (\%) and MRR (\%) are reported. \label{tab:kgqa} }
\begin{tabular}{ccccccc}
	\toprule
Model & Q2B & BetaE & GQE & RotatE & DistMult & ComplEx \\
\hline
MRR & {\bf 51.74} & 50.50 & 36.02 & 43.00 & 38.78 & 41.62\\
Hit@10 & {\bf 68.93} & 64.84 & 52.07 & 55.81 & 49.49 & 53.92 \\
	\bottomrule
\end{tabular}
\end{table}

\section{MRR results of queries with negation}\label{app:negation}
Besides the conjunctive queries and queries with union, we further evaluate the performance of answering the 5 query structures with negation. As the only method that can handle negation, BetaE can effectively model negation especially on ogbl-wikikg2.

\begin{table}[h!]
\centering
\caption{Performance (\%) of BetaE on queries with negation. 
}
\begin{tabular}{cccccccc}
	\toprule
	Dataset & Metrics & 2in & 3in & inp & pin & pni & avg \\
	\hline
	\multirow{2}{*}{FB15k} & MRR & 12.12 & 14.08 & 8.40 & 5.18 & 11.1 & 10.18 \\
		& Hit@10 & 25.45 & 29.64 & 17.75 & 11.11 & 22.97 & 21.38 \\
	\hline
	\multirow{2}{*}{FB15k-237} & MRR & 3.11 & 6.05 & 5.72 & 3.13 & 2.10 & 4.02 \\
		& Hit@10 & 6.82 & 13.36 & 12.72 & 6.69 & 4.08 & 8.734 \\
	\hline
	\multirow{2}{*}{NELL995} & MRR & 3.10 & 5.77 & 7.21 & 2.35 & 2.23 & 4.13 \\
		& Hit@10 & 6.95 & 14.09 & 15.13 & 4.8 & 4.47 & 9.08 \\
	\hline
	\multirow{2}{*}{OGB} & MRR & 35.57 & 50.08 & 38.19 & 38.40 & 37.69 & 39.99 \\
		& Hit@10 & 55.50 & 67.64 & 57.11 & 57.74 & 56.87 & 58.97 \\								
	\bottomrule
\end{tabular}
\end{table}

\section{Baseline code}
Following Marius~\cite{mohoney2021marius}, we adopt the same official release of DGL-KE and PBG. For Marius, we use the release at \url{https://github.com/marius-team/marius/tree/osdi2021}.

\end{document}